\documentclass{article}

\PassOptionsToPackage{numbers}{natbib}

\usepackage[final]{neurips_2020}

\usepackage[utf8]{inputenc} 
\usepackage[T1]{fontenc}    
\usepackage[colorlinks]{hyperref}       
\usepackage{url}            
\usepackage{booktabs}       
\usepackage{amsfonts}       
\usepackage{nicefrac}       
\usepackage{microtype}      

\usepackage{amsmath}
\DeclareMathOperator*{\argmax}{arg\,max}

\usepackage{subfigure}
\usepackage{wrapfig}
\usepackage{multirow}
\usepackage{adjustbox}

\usepackage{color}

\definecolor{BrickRed}{rgb}{0.6,0,0}
\definecolor{RoyalBlue}{rgb}{0,0,0.8}
\definecolor{Tdgreen}{rgb}{0,0.4,0.7}

\hypersetup{citecolor=RoyalBlue}

\newcommand{\sNMargin}{-0.0in}
\newcommand{\ssNMargin}{-0.0in}

\title{Consistency Regularization for Certified Robustness of Smoothed Classifiers}

\author{
  Jongheon Jeong\textsuperscript{\normalfont 1} \qquad Jinwoo Shin\textsuperscript{\normalfont 2,1}
  \\
  \textsuperscript{1}School of Electrical Engineering \quad \textsuperscript{2}Graduate School of AI\\
  Korea Advanced Institute of Science and Technology (KAIST) \\
  Daejeon, South Korea\\
  \texttt{\{jongheonj,\,jinwoos\}@kaist.ac.kr}
}

\begin{document}

\maketitle

\begin{abstract}
  A recent technique of \emph{randomized smoothing} has shown that the worst-case (adversarial) $\ell_2$-robustness can be transformed into the average-case Gaussian-robustness by ``smoothing'' a classifier, i.e., by considering the averaged prediction over Gaussian noise. In this paradigm, one should rethink the notion of adversarial robustness in terms of generalization ability of a classifier under noisy observations. We found that the trade-off between accuracy and certified robustness of smoothed classifiers can be greatly controlled by simply regularizing the prediction consistency over noise. This relationship allows us to design a robust training objective without approximating a non-existing smoothed classifier, e.g., via soft smoothing. Our experiments under various deep neural network architectures and datasets show that the ``certified'' $\ell_2$-robustness can be dramatically improved with the proposed regularization, even achieving better or comparable results to the state-of-the-art approaches with significantly less training costs and hyperparameters.
\end{abstract}

\vspace{\sNMargin}
\section{Introduction}
\label{s:intro}
\vspace{\sNMargin}

Despite achieving even super-human level performance on \textit{i.i.d.}\ datasets \cite{he2016deep, silver2017mastering, devlin-etal-2019-bert}, deep neural network (DNN) classifiers usually make substantially fragile predictions than humans on the samples not from the data-generating distribution. The broad existence of \emph{adversarial examples} \cite{szegedy2013intriguing, goodfellow2014explaining} are arguably the most crucial instance of this phenomenon: a small, adversarially-crafted perturbation on input can easily change the prediction of a classifier, even when the perturbation does not affect the semantic information perceived by humans at all.

This intriguing weakness of DNNs has encouraged many researchers to develop \emph{robust} neural networks, along with a parallel attempt to break them with stronger attacks \cite{carlini2017adversarial, pmlr-v80-uesato18a, pmlr-v80-athalye18a}.
Currently, the community has agreed that \emph{adversarial training} \cite{goodfellow2014explaining, madry2018towards, pmlr-v97-zhang19p}, i.e., augmenting the training dataset with adversarial examples, is an effective defense method, but the ``scalability'' of the method is often questionable in several aspects: (a) it is generally hard to guarantee that an adversarially-trained classifier is indeed robust, (b) generalizing the robustness beyond the training threat model is still challenging \cite{tramer2019adversarial, kang2019testing}, and (c) the network capacity required for robust representation seems to be much larger than practice, e.g., a recent observation shows empirical robustness does not saturate even at ResNet-638 on ImageNet dataset \cite{Xie2020Intriguing}.

Alternatively, a growing body of the research has developed methods that can provide \emph{certified robustness} \cite{sinha2018certifiable, pmlr-v80-wong18a, zhang2020towards}. \emph{Randomized smoothing} \cite{lecuyer2019certified, pmlr-v97-cohen19c} is a recent idea in this direction, which shows that any classifier (e.g., a neural network) that performs well under Gaussian noise can be ``smoothed'' into a certifiably robust classifier. 
This opens a new, scalable notion of adversarial robustness: a neural network may not have to be perfectly smooth, as a proxy of another classifier.

However, it has been relatively under-explored that how to train a good base classifier to maximize the certified robustness of the smoothed counterpart, e.g., \citet{pmlr-v97-cohen19c} only explored the standard training with Gaussian augmentation.
A few recent works \cite{nips_salman19, Zhai2020MACER} have shown that a more sophisticated training algorithm can indeed improve the certified robustness, but the common downside is that they require a sensitive choice of many hyperparameters to optimally trade-off between accuracy and robustness, often imposing a significant amount of additional training costs. 

\textbf{Contribution. }
In this paper, we show that a simple \emph{consistency regularization} term added on a standard training scheme surprisingly improves the certified robustness of smoothed classifiers. Maintaining the prediction consistency over a certain noise, e.g., Gaussian, can be regarded as a natural and desirable property for a classifier under noisy observations. Indeed, for example, forcing such consistency is now considered as one of the most popular techniques in the semi-supervised learning literature \cite{sajjadi2016regularization, miyato2018virtual, oliver2018realistic, berthelot2019mixmatch}. We examine this regularization, motivated by the observation that perfect consistency is a sufficient condition for minimizing the robust 0-1 loss of smoothed classifiers.
This observation connects certified robustness of smoothed classifiers to the general corruption robustness \cite{hendrycks2018benchmarking, pmlr-v97-gilmer19a}, supporting a great potential of smoothed inference as a scalable alternative of adversarially-trained, deterministic classifiers.

We verify the effectiveness of our proposed regularization based on extensive evaluation covering MNIST \cite{dataset/mnist}, CIFAR-10 \cite{dataset/cifar}, and ImageNet \cite{ILSVRC15} classification datasets. 
We show that our simple technique upon a na\"ive training achieves a very comparable, or even better, certified $\ell_2$-robustness to other recent, robust training methods \cite{li2019stab, nips_salman19, Zhai2020MACER}. 
For example, one of our models for CIFAR-10 shows a better robustness than those trained by other tested methods with $2.7\times$ faster training due to its simplicity.
Furthermore, we also demonstrate that applying our method upon a more sophisticated training even further improves the certified robustness, e.g., our method applied upon state-of-the-art training could further improve the average certified $\ell_2$-radius $0.785\rightarrow0.816$ on CIFAR-10.

Despite its effectiveness, our proposed regularization is easy-to-use with fewer hyperparameters, and could run significantly faster than existing approaches without additional backward computation as in adversarial training. We observe that our method does not introduce instability in training for a wide range of hyperparameters, offering a new, stable trade-off term between accuracy and certified robustness of smoothed classifiers. Finally, our concept of regularizing prediction consistency can be extended to other families of noise other than Gaussian, which are often corresponded to different types of adversary, e.g., Laplace noise for $\ell_1$-robustness \cite{lecuyer2019certified, dvijotham2020a, yang2020randomized}.

\vspace{\sNMargin}
\section{Preliminaries}
\label{s:background}
\vspace{\sNMargin}

\vspace{\ssNMargin}
\subsection{Adversarial robustness}
\vspace{\ssNMargin}

We consider a classification task with $K$ classes from a dataset $\mathcal{D}=\{(x_i, y_i)\}^n_{i=1}$,
where $x \in \mathbb{R}^d$ and $y \in \mathcal{Y}:=\{1, \cdots, K\}$ denote an input and the corresponding class label, respectively. Usually, $\mathcal{D}$ is assumed to be \textit{i.i.d.}\ samples from a data-generating distribution $P$.
Let $f: \mathbb{R}^d\rightarrow \mathcal{Y}$ be a classifier.
In many cases, e.g., neural networks, this mapping is modeled by $f(x):=\argmax_{k\in\mathcal{Y}}F_k (x)$ with a differentiable mapping $F: \mathbb{R}^{d}\rightarrow \Delta^{K-1}$ for a gradient-based optimization, where $\Delta^{K-1}$ denotes the probability simplex in $\mathbb{R}^K$. 

In the literature of general robustness research \cite{dodge2017study, biggio2018wild, hendrycks2018benchmarking}, $f$ is required to perform well not only on $P$, but also on a certain extension of it without changing the semantics, say $\widetilde{P}$.
In particular, the notion of \emph{adversarial robustness} considers the worst-case distribution near $P$ under a certain distance metric. More concretely, a common way to define the adversarial robustness is to consider the \emph{average minimum-distance} of adversarial perturbation \cite{moosavi2016deepfool, carlini2017towards, carlini2019evaluating}, namely:
\begin{equation}
\label{eq:avg_min_dist}
    R(f; P) := \mathbb{E}_{(x, y)\sim P}\left[\min_{f (x')\ne y} ||x' - x||_2\right].
\end{equation}
Therefore, our goal is to train $f$ that (a) performs well on $P$, while (b) maximizing $R(f; P)$ as well.

\vspace{\ssNMargin}
\subsection{Randomized smoothing}
\vspace{\ssNMargin}

In practice, the inner minimization objective in \eqref{eq:avg_min_dist} is usually not easy to optimize exactly, and mostly results in near-zero value on standard neural network classifiers. The key idea of \emph{randomized smoothing} \cite{pmlr-v97-cohen19c} is rather to consider the robustness of a ``smoothed'' transformation of the \emph{base classifier} $f$ over Gaussian noise, namely $\hat{f}$:
\begin{equation}
\label{eq:smoothing}
    \hat{f}(x) := \argmax_{k\in \mathcal{Y}} \mathbb{E}_{\delta\sim\mathcal{N}(0, \sigma^2 I)}\left[\mathbf{1}_{f (x + \delta)= k} \right],
\end{equation}
where $\mathbf{1}_{A}$ denotes the \emph{indicator} random variable, formally defined by $\mathbf{1}_{A}(\omega)=1$ if $\omega \in A$ and $0$ otherwise, and
$\sigma^2$ is a hyperparameter that controls the level of smoothing. 
For a given input $x$, \citet{pmlr-v97-cohen19c} guarantees a \emph{certified radius} in $\ell_2$ distance, the current state-of-the-art lower bound of the minimum-distance of adversarial perturbation around $\hat{f}(x)$:
suppose $f(x+\delta)$ returns a class $\hat{f}(x)\in\mathcal{Y}$ with probability $p^{(1)}$ and the ``runner-up'' (i.e., the second best) class with probability $p^{(2)}:=\max_{c\ne \hat{f}(x)}\mathbb{P}(f(x+\delta)=c)$. Then, the lower bound can be given as follows: 
\begin{equation}
\label{eq:cr}
    R(\hat{f}; x, y) := \min_{\hat{f} (x')\ne y} ||x' - x||_2 \ge \frac{\sigma}{2}\left(\Phi^{-1}(p^{(1)}) - \Phi^{-1}(p^{(2)}) \right) 
\end{equation}
provided that $\hat{f}(x) = y$, otherwise $R(\hat{f}; x, y) := 0$. Here, $\Phi$ denotes the standard Gaussian CDF.
Since the inequality holds for any upper bound of $p^{(2)}$, say $\overline{p^{(2)}}$, one could also obtain a bit loose, but simpler bound of certified radius by letting $\overline{p^{(2)}} = 1 - p^{(1)} \ge p^{(2)}$:
\begin{equation}
\label{eq:lower_bound}
    R(\hat{f}; x, y) \ge \sigma\cdot\Phi^{-1}(p^{(1)}) =: \underline{R}(\hat{f}, x, y).
\end{equation}

\vspace{\sNMargin}
\section{Consistency regularization for smoothed classifiers}
\label{s:method}
\vspace{\sNMargin}

Our intuition on the proposed consistency regularization is based on minimizing the \emph{0-1 robust classification loss}, in a similar manner to the recent attempts of decomposing the training objective with respect to the accuracy and robustness \cite{pmlr-v97-zhang19p, Zhai2020MACER}. Specifically, we attempt minimize the following:
\begin{equation}
\label{eq:decomp}
   \mathbb{E}_{(x, y)\in\mathcal{D}}\left[1 - \mathbf{1}_{\underline{R}(\hat{f}; x, y) \ge \varepsilon}\right]
    = \underbrace{\mathbb{E}\left[\mathbf{1}_{\hat{f}(x) \ne y}\right]}_{\text{natural error}} + \underbrace{\mathbb{E}\left[\mathbf{1}_{\hat{f}(x) = y,\ \underline{R}(\hat{f}; x, y) < \varepsilon}\right]}_{\text{robust error}},
\end{equation}
where $\underline{R}$ is the certified lower bound of $R$ as defined in \eqref{eq:lower_bound}, and 
$\varepsilon > 0$ is a pre-defined constant.
Assuming that the natural error term can be optimized via a standard surrogate loss, e.g., cross-entropy, we rather focus on how to minimize the robust error term. Here, the key difficulties to consider a gradient-based optimization is that (a) computing $\hat{f}$ exactly is intractable, and more importantly, (b) $\hat{f}$ is practically a non-differentiable object when estimated via Monte Carlo sampling (see \eqref{eq:smoothing}), so that even a proper surrogate loss function would not make the optimization differentiable.

To bypass these issues, we instead concentrate on a \emph{sufficient condition} to minimize the given robust 0-1 loss. Recall that we assume $f(x)=\argmax_{k\in\mathcal{Y}} F(x)$ for a differentiable function $F: \mathbb{R}^{d}\rightarrow \Delta^{K-1}$.
Here, we notice that the robust loss in \eqref{eq:decomp} would anyway become zero if $F(x+\delta)$ returns a \emph{constant} output over $\delta$ for a given $x$. Indeed, this implies $\mathbb{P}_{\delta}(f(x+\delta)=\hat{f}(x))$ to become 1 \emph{regardless} of what $\hat{f}$ is, and minimizes an upper bound of the robust loss in \eqref{eq:decomp} due to the following:
\begin{align}\label{eq:rob_obj}
    \mathbb{E}_{(x, y)\in\mathcal{D}}\left[\mathbf{1}_{\hat{f}(x) = y,\ \underline{R}(\hat{f}; x, y) < \varepsilon}\right]
    &= \mathbb{E}\left[\mathbf{1}_{\hat{f}(x) = y,\ \underline{R}(\hat{f}; x, \hat{f}(x)) < \varepsilon}\right] \nonumber \\
    &\le \mathbb{E}\left[\mathbf{1}_{\underline{R}(\hat{f}; x, \hat{f}(x)) < \varepsilon}\right] = \mathbb{E}\left[\mathbf{1}_{\mathbb{P}_\delta(f(x+\delta)=\hat{f}(x)) < \Phi(\frac{\varepsilon}{\sigma})}\right],
\end{align}
where the last equality is from the definition of $\underline{R}$ in \eqref{eq:lower_bound}.
Therefore, we attempt to optimize the robust training objective on $\hat{f}$ via regularizing $F(x+\delta)$ to be \emph{consistent} across $\delta$. Specifically, we propose the following consistency regularization upon any standard training objective:
\begin{equation}
\label{eq:reg}
    L^{\tt con} := \lambda \cdot \mathbb{E}_\delta\left[\mathrm{KL}(\hat{F}(x) || F(x + \delta))\right] 
    + \eta \cdot \mathrm{H}(\hat{F}(x)),
\end{equation}
where $\hat{F}(x):=\mathbb{E}[F(x+\delta)]$ is the mean of $F(x+\delta)$, $\mathrm{KL}(\cdot||\cdot)$ and $\mathrm{H}(\cdot)$ denote the Kullback–Leibler (KL) divergence and the entropy, respectively, and $\lambda, \eta > 0$ are hyperparameters that control the relative strength.
In other words, this regularization enforces $F$, correspondingly $f$ as well, to reduce the \emph{variance} of predictions under Gaussian noise for a given sample $x$, while preventing the mean to be too close to the uniform via the entropy penalty.
Note that the proposed form \eqref{eq:reg} includes the cross-entropy loss $\mathbb{E}[\mathcal{L}(F(x+\delta), \hat{F}(x))]$ when $\lambda=\eta$. In practice, we observe $\lambda$ plays a more crucial role than $\eta$ for the trade-off between accuracy and robustness: e.g., in our experiments, we use a fixed $\eta=0.5$ unless otherwise noted, and adjust $\lambda$ to control the robustness.

{We also remark that, for a fixed $x$, \eqref{eq:reg} gives a family of \emph{calibrated} \cite{pires2016multiclass, pmlr-v97-zhang19p} surrogate losses of the 0-1 risk $\mathbb{E}_{\delta}[\mathbf{1}_{f(x+\delta)\neq \hat{f}(x)}]=\mathbb{P}_\delta(f(x+\delta)\neq \hat{f}(x))$, i.e., minimizers of \eqref{eq:reg} are also those of the 0-1 risk, and {thus it} minimizes the upper bound in \eqref{eq:rob_obj} when minimized across $(x, y)\sim\mathcal{D}$.} 
Similarly, one can adopt other forms of consistency regularization as long as the regularization leads \eqref{eq:rob_obj} to be zero: we examine such variants in Section~\ref{ss:ablation}, and it turns out indeed they are also effective to improve the certified robustness, while our form \eqref{eq:reg} shows a particular robustness compared to them empirically.

\begin{figure}
	\centering
	\hfill
	\subfigure[SmoothAdv \cite{nips_salman19}]
	{
	    \includegraphics[width=0.31\linewidth]{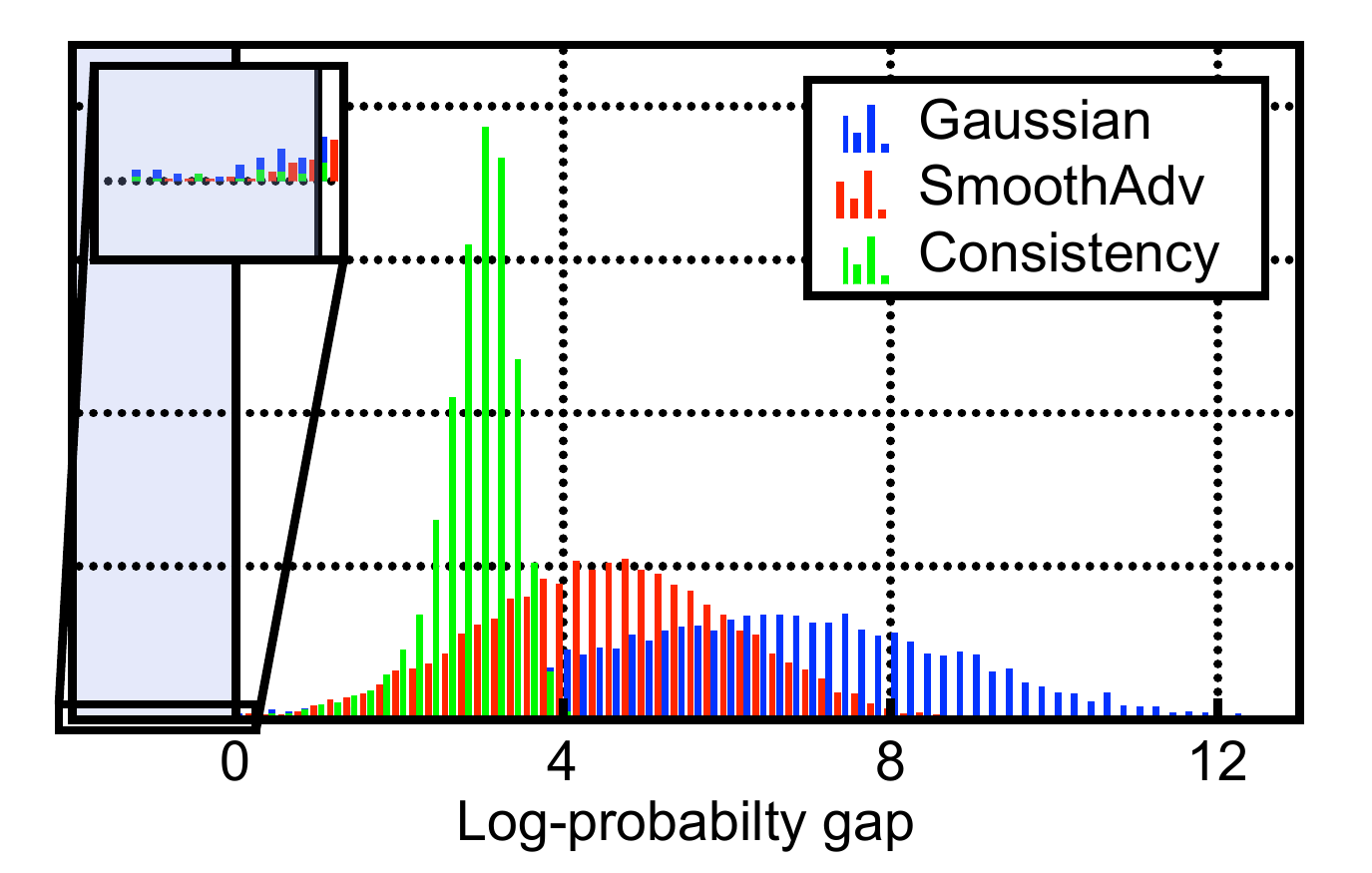}
		\label{fig:hist_main}
	}
	\hfill
	\subfigure[MACER \cite{Zhai2020MACER}]
	{
	    \includegraphics[width=0.31\linewidth]{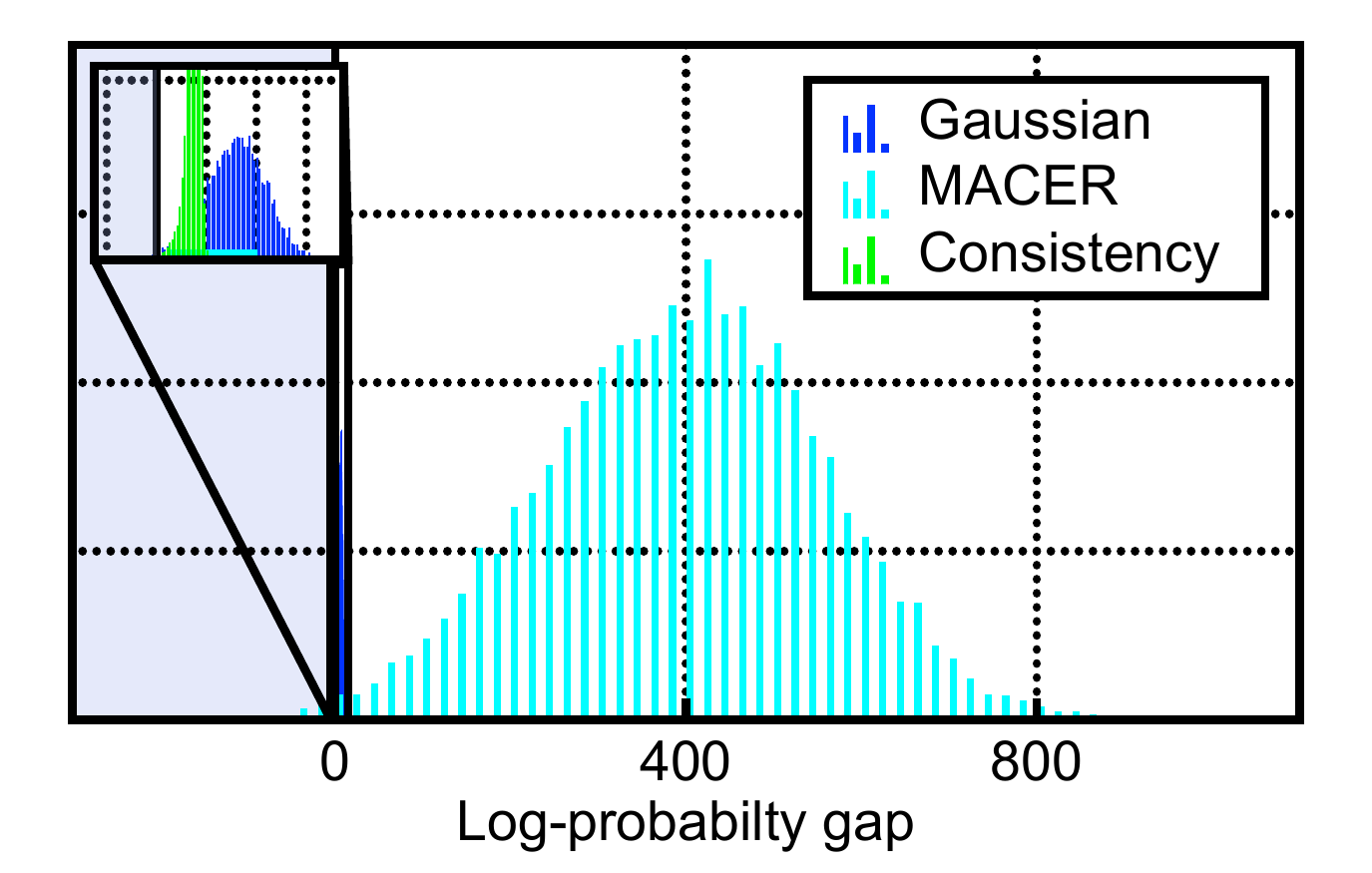}
		\label{fig:hist_macer}
	}
	\hfill
	\subfigure[Stability training \cite{li2019stab}]
	{
	    \includegraphics[width=0.31\linewidth]{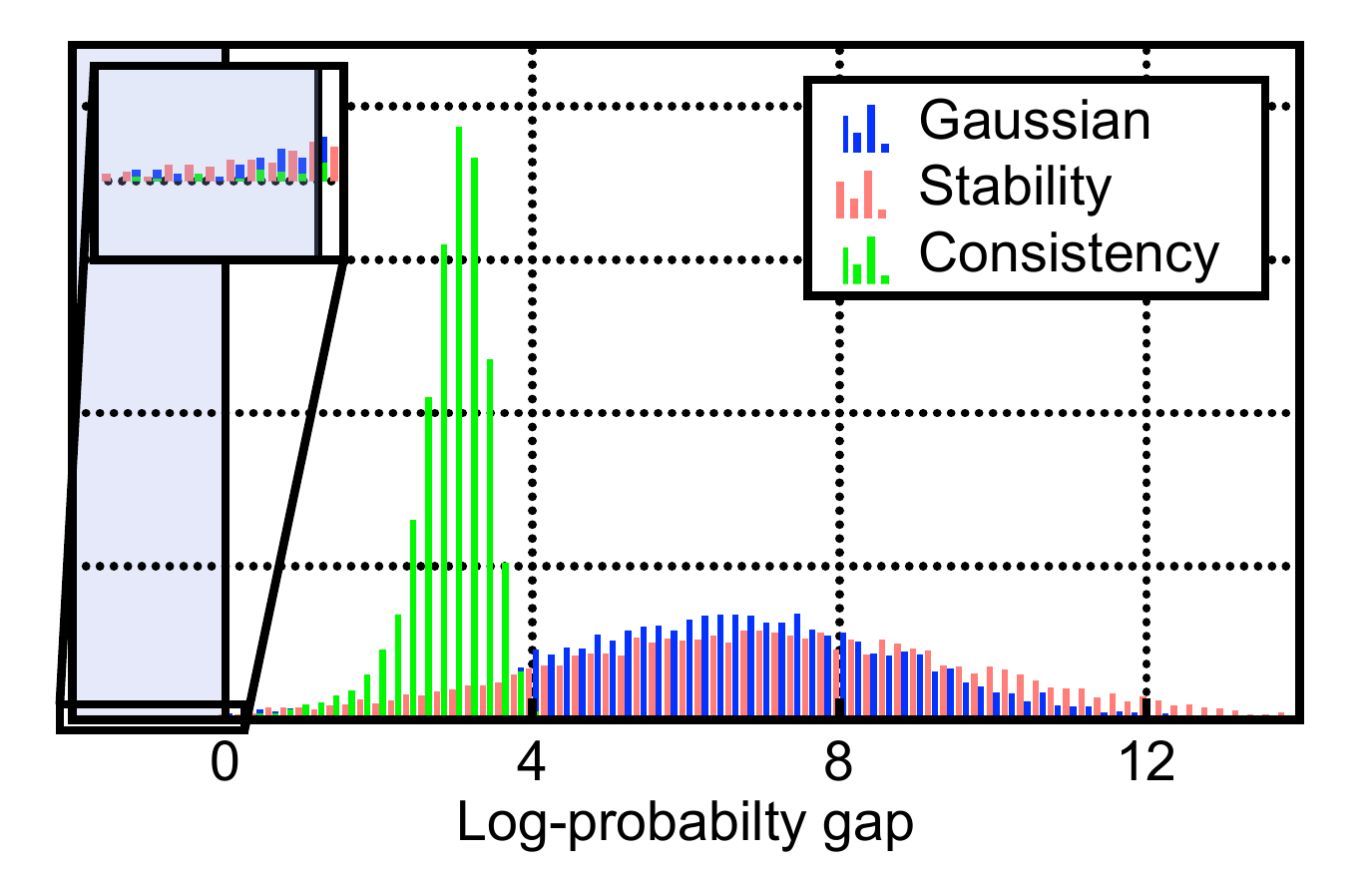}
		\label{fig:hist_stab}
	}
	\hfill
	\vspace{-0.05in}
	\caption{Comparison of log-probabilty  distributions under Gaussian noise at a fixed test sample of MNIST. For each histogram, we use 10,000 samples of noise. {``Gaussian'' indicates the baseline training with Gaussian augmentation \cite{pmlr-v97-cohen19c}, and ``Consistency'' indicates our proposed regularization applied upon ``Gaussian''.} The left, shaded areas are where a classifier makes a misclassification.}
	\label{fig:logprob}
	\vspace{-0.1in}
\end{figure}

\vspace{\ssNMargin}
\subsection{Comparison to prior works}
\vspace{\ssNMargin}

There have been a few prior approaches in attempts to improve the robustness of smoothed classifiers with a more sophisticated training method beyond that of \citet{pmlr-v97-cohen19c}. \citet{nips_salman19} proposed \emph{SmoothAdv}, which shows that adversarial training directly on smoothed classifiers improve the certified robustness. More recently, \citet{Zhai2020MACER} proposed a faster training method called \emph{MACER}, via maximizing a soft approximation of the certified radius given in \eqref{eq:cr}.\footnote{For the interested readers, we present a detailed overview of prior works in the supplementary material.} The essential difference of our regularization to the previous works is at how the non-differentiable $\hat{f}$ is handled, namely,  
prior works commonly approximate $\hat{f}$ directly by the inner soft-classifier $F$:
\begin{equation}\label{eq:previous_approx}
    \mathbb{E}_{\delta}[\mathbf{1}_{f (x + \delta)= k}] \approx \mathbb{E}_{\delta}[F_k(x+\delta)],
\end{equation}
for each class $k\in\mathcal{Y}$. 
A key caveat here is that, however, optimizing $\hat{f}$ with this approximation would implicitly count out much optimal solutions of $F$. More specifically, we remark that an optimal soft classifier $F$ does not require to have confidence near to 1 for maximizing the certified radius \eqref{eq:cr}, which is a usual solution found by minimizing the cross-entropy based on \eqref{eq:previous_approx}. 
Our approach rather considers an ``indirect'' regularizer of $\hat{f}$ without assuming such an approximation, thereby allows a more flexible optimization. 

On the other hand, \citet{li2019stab} proposed \emph{stability training}, as a parallel attempt to the Gaussian training of randomized smoothing \cite{pmlr-v97-cohen19c} to obtain a robust smoothed classifier: namely, in order to perform well on Gaussian noise, stability training trains $F$ with the following loss:
\begin{equation}\label{eq:stab}
    \min_F{\mathcal{L}(F(x), y) + \lambda \cdot \mathcal{L}(F(x), F(x+\delta))},
\end{equation}
where $\mathcal{L}$ is the cross-entropy loss. The regularization term used in \eqref{eq:stab} has a seemingly similar formula to ours particularly when $\lambda=\eta$ in \eqref{eq:reg}, but there is a fundamental difference: our method \eqref{eq:reg} does not require $F$ to 
minimize $\mathcal{L}(F(x), y)$ to perform well on $(x+\delta, y)$.
Consequently, our method again allows a more flexible solution compared to \eqref{eq:stab}.
In Section~\ref{ss:cifar10}, we empirically show that our form of regularization \eqref{eq:reg} attains a significantly better robustness than \eqref{eq:stab}. We also show in the supplementary material that, due to such flexibility, our method is much more robust on the choice of $\lambda$.

{\textbf{Log-probability gap. }} In Figure~\ref{fig:logprob}, we illustrate how the optimal classifier found by our method differs from others, by comparing the distribution of \emph{log-probability gap} over Gaussian noise $\delta$ for a given (noisy) sample $(x+\delta, y)$, namely $\log F_y(x+\delta) - \max_{c\ne y} \log F_c(x+\delta)$. 
{In other words, we compare the \emph{output margin} of $f$ at $(x+\delta, y)$ to observe the \emph{input margin} of $\hat{f}$: the robustness guarantee in \eqref{eq:cr} implies that it is enough to minimize $\mathbb{P}_\delta(f(x+\delta)\neq y)$ to improve the robustness of $\hat{f}$ at $x$. This can be also viewed under the Lipschitzness angle: \citet{nips_salman19} show that any Gaussian-smoothed classifier $\hat{f}$ has an explicit Lipschitz constant, leading to a simpler proof of \eqref{eq:cr}.}

Overall, we observe in Figure~\ref{fig:logprob} that our method learns relatively lower, yet more consistent, confidences than others. We also found that MACER \cite{Zhai2020MACER} tends to vary on much larger values in logits (see Figure~\ref{fig:hist_macer}): 
MACER essentially maximizes the gap between the first- and second-best logits of $\mathbb{E}[F(x+\delta)]$, which leads $F$ to have an arbitrary large value when optimized.
Finally, Figure~\ref{fig:hist_stab} supports that our method is fundamentally different to the stability training \cite{li2019stab}: one can observe that stability training does not give a particular consistency that our method shows.

\vspace{\ssNMargin}
\subsection{Training with consistency regularization}
\vspace{\ssNMargin}

\textbf{Overall training objective. }
Combining our regularization $L^{\tt con}$ to a natural surrogate loss $L^{\tt nat}$
leads to a full objective to minimize. Any form of $L^{\tt nat}$ is possible to use, as long as it minimizes the natural error of $\hat{f}$ in \eqref{eq:decomp}, e.g., the standard cross-entropy loss on $f$ may not be proper for $L^{\tt nat}$.
As a plain example, we use the loss proposed by \citet{pmlr-v97-cohen19c}, which simply performs Gaussian augmentation during training: 
for a given sample $(x, y)\sim\mathcal{D}$, the authors suggest to minimize:
\begin{equation}
\label{eq:cohen}
    L^{\tt nat} := \mathbb{E}_{\delta\sim\mathcal{N}(0, \sigma^2 I)}\left[\mathcal{L}(F(x+\delta), y)\right].
\end{equation}
The overall objective with consistency regularization is then:
\begin{align}\label{eq:overall_detailed}
    L &:= L^{\tt nat} + L^{\tt con} = \mathbb{E}_{\delta}\left[
    \mathcal{L}(F(x+\delta), y) + \lambda \cdot \mathrm{KL}(\hat{F}(x) || F(x + \delta)) + \eta \cdot \mathrm{H}(\hat{F}(x))
    \right] \\
    &\ \approx \frac{1}{m}\sum_i \left(\mathcal{L}(F(x+\delta_i), y) + \lambda \cdot \mathrm{KL}(\hat{F}(x) || F(x + \delta_i))\right) + \eta \cdot \mathrm{H}(\hat{F}(x)),
    \label{eq:overall_detailed_mcmc}
\end{align}
where \eqref{eq:overall_detailed_mcmc} is a concrete loss of \eqref{eq:overall_detailed} via Monte Carlo sampling over $\delta$.
Nevertheless, our regularization scheme is not limited to a specific choice of $L^{\tt nat}$, and one can also apply others in a similar way. In our experiments, for example, we show that using SmoothAdv \cite{nips_salman19} as $L^{\tt nat}$ could further improve the certified robustness, although it is significantly more expensive to optimize compared to \eqref{eq:cohen}.

\textbf{Computational overhead. }
We use $m$ independent samples of Gaussian noise in \eqref{eq:overall_detailed_mcmc} to estimate \eqref{eq:overall_detailed}, and this is the only source of extra training costs compared to the original \cite{pmlr-v97-cohen19c}. 
Nevertheless, we empirically observe that the minimal choice\footnote{{We remark the regularization term in \eqref{eq:overall_detailed_mcmc} requires $m>1$ to work, as the KL term in \eqref{eq:reg} would vanish when $m=1$: with only a single sample, say $\delta_1$, $F(x+\delta_1)$ would be the best estimation of $\hat{F}(x)$.}} of $m=2$ is fairly enough for our method, as demonstrated in Figure~\ref{fig:ab/effect_m}.
Considering that other existing methods also use this inner-sampling procedure, often requiring an additional outer-loop of backward computations for adversarial training \cite{nips_salman19}, or a large number of $m$ for a stable training \cite{Zhai2020MACER}, our method offers a significantly less training cost with less hyperparameters, as further discussed in Section~\ref{ss:runtime}.

\vspace{\sNMargin}
\section{Experiments}
\label{s:experiments}
\vspace{\sNMargin}

We validate the effectiveness of our proposed consistency regularization for a wide range of image classification datasets: MNIST\footnote{{All the results on MNIST are provided in the supplementary material.}} \cite{dataset/mnist}, CIFAR-10 \cite{dataset/cifar}, and ImageNet \cite {ILSVRC15}.\footnote{Code is available at \url{https://github.com/jh-jeong/smoothing-consistency}.}
Overall, our results consistently demonstrate that simply applying our method in addition to the other baseline training methods greatly boosts the certified $\ell_2$-robustness via randomized smoothing. Remarkably, we show that our method even further improves the previous state-of-the-art results of SmoothAdv \cite{nips_salman19}.
We also perform an ablation study to further investigate the detailed components proposed in our method.

\begin{figure}[t]
	\centering
	\subfigure[$\sigma=0.25$]
	{
	    \includegraphics[width=0.31\linewidth]{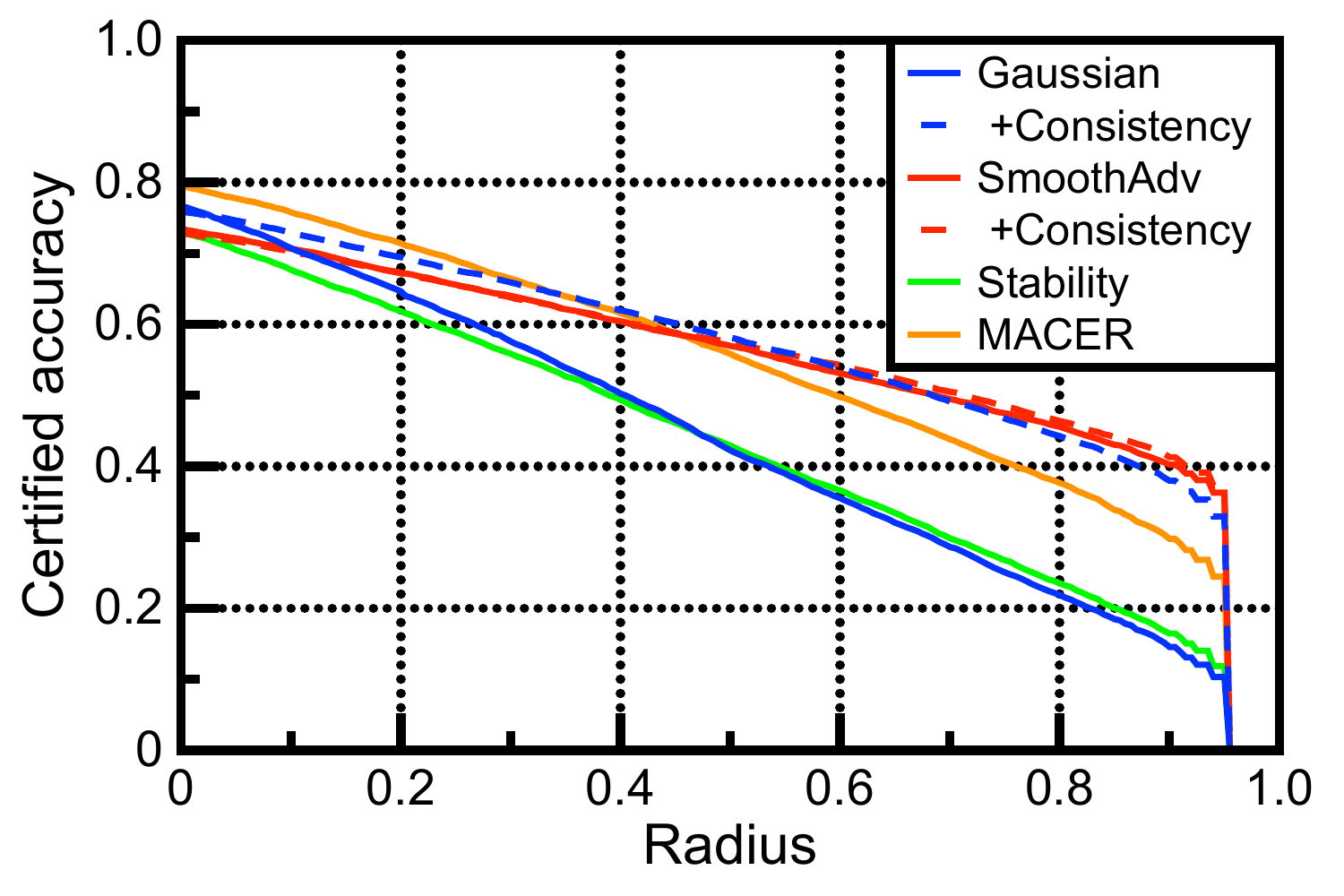}
		\label{fig:cifar10_25}
	}
	\subfigure[$\sigma=0.50$]
	{
	    \includegraphics[width=0.31\linewidth]{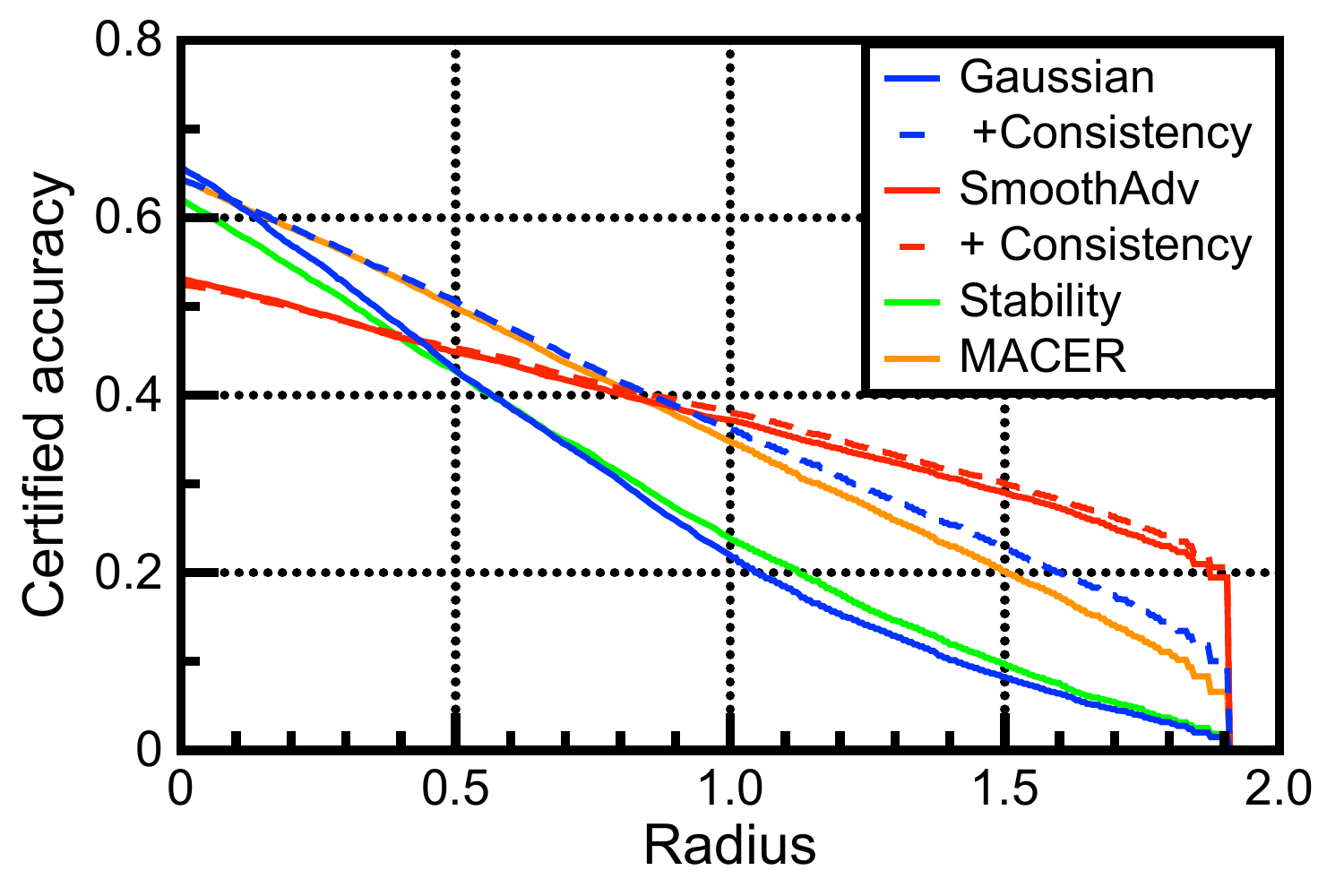}
		\label{fig:cifar10_50}
	}
	\subfigure[$\sigma=1.00$]
	{
	    \includegraphics[width=0.31\linewidth]{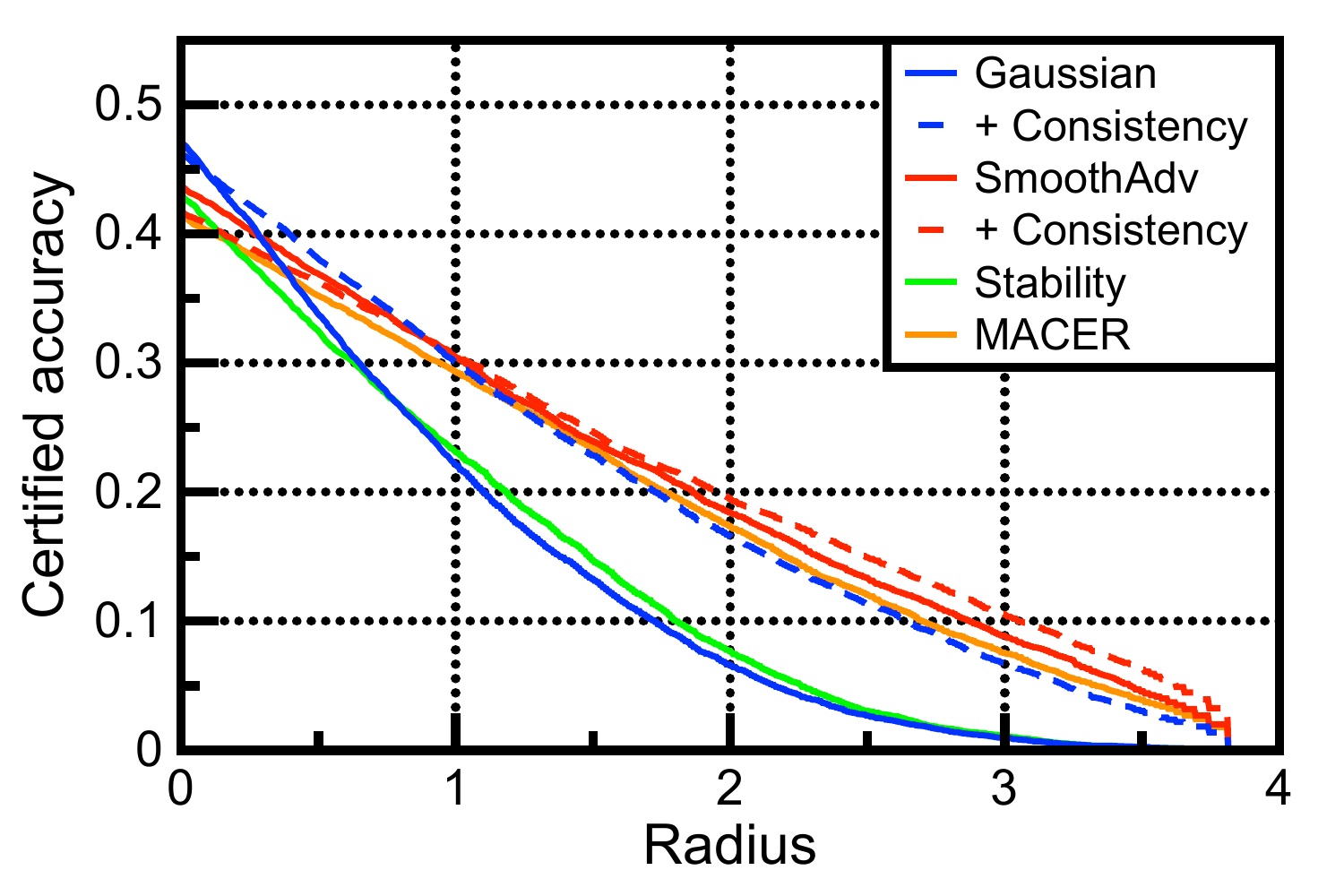}
		\label{fig:cifar10_100}
	}
	\vspace{-0.05in}
	\caption{Comparison of approximate certified accuracy via randomized smoothing for various training methods on CIFAR-10. A sharp drop of certified accuracy in the plots exists since there is a hard upper bound that \textsc{Certify} can output for a given $\sigma$ and $n=100,000$.}
	\label{fig:cifar10}
	\vspace{-0.1in}
\end{figure}

\vspace{\ssNMargin}
\subsection{Setups}
\vspace{\ssNMargin}

\textbf{Evaluation metrics. }
To evaluate certified robustness for a given classifier $f$, we aim to compute the \emph{certified test accuracy} at radius $r$, which is defined by the fraction of the test dataset that $\hat{f}$ can certify the robustness of radius $r$ with respect to the certifiable lower bound in \eqref{eq:cr}.
Due to the intractability of this metric, however, we instead measure the \emph{approximate certified test accuracy} \cite{pmlr-v97-cohen19c}. More concretely, \citet{pmlr-v97-cohen19c} proposed a practical Monte Carlo based certification procedure, namely \textsc{Certify}, which returns the prediction of $\hat{f}$ and a ``safe'' lower bound of certified radius over the randomness of $n$ samples with probability at least $1-\alpha$, or abstains the certification. The \emph{approximate certified test accuracy} is then defined by the fraction of the test dataset which \textsc{Certify} classifies correctly with radius larger than $r$ without abstaining. 

In our experiments, we use the official implementation\footnote{\url{https://github.com/locuslab/smoothing}} of \textsc{Certify} for evaluation, with $n=100,000$, $n_0=100$ and $\alpha=0.001$, following prior works \cite{pmlr-v97-cohen19c, nips_salman19}. We mainly report the approximate certified test accuracy at various radii, but also report the \emph{average certified radius} (ACR) considered by \citet{Zhai2020MACER}, i.e., the averaged value of certified radii returned by \textsc{Certify}, as another metric for better comparison of robustness under the trade-offs between accuracy and robustness \cite{tsipras2018robustness, pmlr-v97-zhang19p}, namely $\mathrm{ACR} := \frac{1}{|\mathcal{D}_{\tt test}|}\sum_{(x, y)\in\mathcal{D}_{\tt test}}\mathrm{CR}(f, \sigma, x)\cdot \mathbf{1}_{\hat{f}(x)=y}$,
where $\mathcal{D}_{\tt test}$ is the test dataset, and $\mathrm{CR}$ denotes the certified radius returned from $\textsc{Certify}(f, \sigma, x)$.

\textbf{Training details. }
We use the same base classifier used in the prior work \cite{pmlr-v97-cohen19c, nips_salman19, Zhai2020MACER}: namely, we use LeNet \cite{dataset/mnist} for MNIST, ResNet-110 \cite{he2016deep} for CIFAR-10, and ResNet-50 \cite{he2016deep} for ImageNet. For a fair comparison, we follow the same training details used in \citet{pmlr-v97-cohen19c} and \citet{nips_salman19}. For each model configuration, we consider three different models as varying the noise level $\sigma\in\{0.25, 0.5, 1.0\}$. During inference, we apply randomized smoothing with the same $\sigma$ used in the training. When our regularization is used, we use $m=2$ and $\eta=0.5$ unless otherwise specified. More training details are specified in the supplementary material.

\textbf{Baseline methods. }
We evaluate how consistency regularization would affect the certified robustness when applied to a baseline training method. In our experiments, we consider two baseline methods proposed for training smoothed classifiers to apply our regularization scheme: (a) Gaussian \cite{pmlr-v97-cohen19c}: training with Gaussian augmentation over $\mathcal{N}(0, \sigma^2 I)$; (b) SmoothAdv \cite{nips_salman19}: adversarial training on a soft approximation of the smoothed classifier. We also consider {stability training \cite{li2019stab} in \eqref{eq:stab} and MACER \cite{Zhai2020MACER} to compare, as other regularization-based approaches.}

\begin{table}[t]
\centering
\caption{Comparison of approximate certified test accuracy (\%) on CIFAR-10. Every model is certified with $\sigma$ used for its training. We set our result bold-faced whenever the value improves the baseline. For ACRs, we underline the best model per $\sigma$. For the results in ``+ Hyperparameter search'', we evaluate the best model among those released by \citet{nips_salman19} for each $\sigma$.}
\label{tab:cifar10}
    \vspace{-0.02in}
    \begin{adjustbox}{width=1\linewidth}
    \begin{tabular}{clccccccccccc}
    \toprule
    $\sigma$ &  Models (CIFAR-10) & ACR & 0.00 & 0.25 & 0.50 & 0.75 & 1.00 & 1.25 & 1.50 & 1.75 & 2.00 & 2.25 \\ 
    \midrule
    \multirow{8}{*}{0.25}& Gaussian \cite{pmlr-v97-cohen19c} & {0.424} & {76.6} & {61.2} & {42.2} & {25.1} & 0.0 & 0.0 & 0.0 & 0.0 & 0.0 & 0.0  \\
    & \textbf{+ Consistency ($\lambda=10$)} & \textbf{{0.544}}  & \textbf{77.8} & \textbf{68.8} & \textbf{57.4} & \textbf{43.8} & 0.0 & 0.0 & 0.0 & 0.0 & 0.0 & 0.0  \\
    & \textbf{+ Consistency ($\lambda=20$)} & \textbf{\underline{0.552}}  & 75.8 & \textbf{67.6} & \textbf{58.1} & \textbf{46.7} & 0.0 & 0.0 & 0.0 & 0.0 & 0.0 & 0.0  \\
    \cmidrule(l){2-2} \cmidrule(l){3-3} \cmidrule(l){4-13}
    & SmoothAdv \cite{nips_salman19} & {0.544} & {73.4} & {65.6} & {57.0} & {47.5} & 0.0 & 0.0 & 0.0 & 0.0 & 0.0 & 0.0  \\
    & \textbf{+ Consistency ($\lambda=2$)} & \textbf{0.548} & 72.9 & 65.6 & \textbf{57.5} & \textbf{48.5} & 0.0 & 0.0 & 0.0 & 0.0 & 0.0 & 0.0 \\ 
    \cmidrule(l){2-2} \cmidrule(l){3-3} \cmidrule(l){4-13}
    & Stability training \cite{li2019stab} & 0.421 & 72.3 & 58.0 & 43.3 & 27.3 & 0.0 & 0.0 & 0.0 & 0.0 & 0.0 & 0.0  \\ 
    & MACER  \cite{Zhai2020MACER} & {0.531}  & {79.5} & {69.0} & {55.8} & {40.6} & 0.0 & 0.0 & 0.0 & 0.0 & 0.0 & 0.0 \\  \midrule
    \multirow{8}{*}{0.50}& Gaussian \cite{pmlr-v97-cohen19c} & {0.525} & {65.7} & {54.9} & {42.8} & {32.5} & {22.0} & {14.1} & {8.3} & {3.9} & 0.0 & 0.0  \\
    & \textbf{+ Consistency ($\lambda=10$)}  & {\textbf{0.720}}  & 64.3 & \textbf{57.5} & \textbf{50.6} & \textbf{43.2} & \textbf{36.2} & \textbf{29.5} & \textbf{22.8} & \textbf{16.1} & 0.0 & 0.0 \\
    \cmidrule(l){2-2} \cmidrule(l){3-3} \cmidrule(l){4-13}
    & SmoothAdv \cite{nips_salman19} & 0.689 & 64.4 & 57.2 & 49.0 & 40.6 & 33.6 & 27.4 & 21.8 & 14.0 & 0.0 & 0.0 \\
    & + Hyperparameter search  & {0.717} & {53.1} & {49.2} & {44.9} & {41.0} & {37.2} & {33.2} & {29.1} & {24.0} & 0.0 & 0.0 \\  
    & \textbf{+ Consistency ($\lambda=1$)} & \textbf{\underline{0.726}} & 52.3 & 48.9 & \textbf{45.1} & \textbf{41.3} & \textbf{37.8} & \textbf{33.9} & \textbf{29.9} & \textbf{25.2} & 0.0 & 0.0  \\ 
    \cmidrule(l){2-2} \cmidrule(l){3-3} \cmidrule(l){4-13}
    & Stability training \cite{li2019stab} & {0.521} & 60.6 & 51.5 & 41.4 & 32.5 & 23.9 & 15.3 & 9.6 & 5.0 & 0.0 & 0.0 \\ 
    & MACER \cite{Zhai2020MACER} & {0.691} & {64.2} & {57.5} & {49.9} & {42.3} & {34.8} & {27.6} & {20.2} & {12.6} & 0.0 & 0.0  \\ 
    \midrule
    \multirow{9}{*}{1.00}& Gaussian \cite{pmlr-v97-cohen19c} & 0.542 & 47.2 & 39.2 & 34.0 & 27.8 & 21.6 & 17.4 & 14.0 & 11.8 & 10.0 & 7.6 \\
    & \textbf{+ Consistency ($\lambda=5$)}  & {\textbf{0.734}} & \textbf{48.1} & \textbf{43.9} & \textbf{39.3} & \textbf{34.7} & \textbf{29.9} & \textbf{26.1} & \textbf{22.1} & \textbf{18.8} & \textbf{15.4} & \textbf{12.2} \\
    & \textbf{+ Consistency ($\lambda=10$)}  & {\textbf{0.756}} & 46.3 & \textbf{42.2} & \textbf{38.1} & \textbf{34.3} & \textbf{30.0} & \textbf{26.3} & \textbf{22.9} & \textbf{19.7} & \textbf{16.6} & \textbf{13.8} \\
    \cmidrule(l){2-2} \cmidrule(l){3-3} \cmidrule(l){4-13}
    & SmoothAdv \cite{nips_salman19} & 0.682 & 50.2 & 44.0 & 37.6 & 33.8 & 28.8 & 24.0 & 20.2 & 15.8 & 13.2 & 10.2 \\
    & + Hyperparameter search & {0.785} & {45.6} & {41.9} & {38.0} & {34.2} & {30.9} & {27.4} & {24.1} & {20.7} & {17.7} & {14.9}  \\ 
    & \textbf{+ Consistency ($\lambda=1$)} & {\underline{\textbf{0.816}}} & 41.7 & 39.0 & 36.2 & 33.5 & 30.7 & \textbf{27.6} & \textbf{24.7} & \textbf{22.0} & \textbf{19.5} & \textbf{17.3}  \\ 
    \cmidrule(l){2-2} \cmidrule(l){3-3} \cmidrule(l){4-13}
    & Stability training \cite{li2019stab} & 0.526  & 43.5 & 38.9 & 32.8 & 27.0 & 23.1 & 19.1 & 15.4 & 11.3 & 7.8 & 5.7 \\
    & MACER  \cite{Zhai2020MACER} & {0.744} & {41.4} & {38.5} & {35.2} & {32.3} & {29.3} & {26.4} & {23.4} & {20.2} & {17.4} & {14.5}  \\ 
    \bottomrule
    \end{tabular}
    \end{adjustbox}
    \vspace{-0.05in}
\end{table}
\begin{table}
\centering
\caption{Comparison of approximate certified test accuracy (\%) on ImageNet. We set our result bold-faced whenever the value improves the baseline. We use $\eta=0.1$ instead of $0.5$ when $\sigma=1.0$.}
\label{tab:imagenet}
\vspace{-0.02in}
\small
    \begin{tabular}{clccccccccc}
    \toprule
    $\sigma$ &  Models (ImageNet) & ACR & 0.0 & 0.5 & 1.0 & 1.5 & 2.0 & 2.5 & 3.0 & 3.5 \\
    \midrule
    \multirow{3.5}{*}{0.50}& Gaussian \cite{pmlr-v97-cohen19c} & 0.733 & 57 & 46 & 37 & 29 & 0 & 0 & 0 & 0  \\
    & \textbf{+ Consistency ($\lambda=5$)}  & \textbf{0.822} & 55 & \textbf{50} & \textbf{44} & \textbf{34} & 0 & 0 & 0 & 0 \\
    \cmidrule(l){2-2} \cmidrule(l){3-3} \cmidrule(l){4-11}
    & SmoothAdv \cite{nips_salman19} & 0.825 & 54 & 49 & 43 & 37 & 0 & 0 & 0 & 0  \\
    \midrule
    \multirow{3.5}{*}{1.00}& Gaussian \cite{pmlr-v97-cohen19c} & 0.875 & 44 & 38 & 33 & 26 & 19 & 15 & 12 & 9  \\
    & \textbf{+ Consistency ($\lambda=5$)}  & \textbf{0.982} & 41 & 37 & 32 & \textbf{28} & \textbf{24} & \textbf{21} & \textbf{17} & \textbf{14} \\
    \cmidrule(l){2-2} \cmidrule(l){3-3} \cmidrule(l){4-11}
    & SmoothAdv \cite{nips_salman19} & 1.040 & 40 & 37 & 34 & 30 & 27 & 25 & 20 & 15 \\
    \bottomrule
    \end{tabular}
    \vspace{-0.1in}
\end{table}

\vspace{\ssNMargin}
\subsection{Results on CIFAR-10}
\label{ss:cifar10}
\vspace{\ssNMargin}

We train CIFAR-10 models for 150 epochs following the training details of SmoothAdv \cite{nips_salman19}. 
Whenever possible, we use the pre-trained models officially released by the authors for our evaluation to reproduce the baseline results.\footnote{\url{https://github.com/Hadisalman/smoothing-adversarial}}\footnote{\url{https://github.com/RuntianZ/macer}}\footnote{{In case of the pre-trained MACER models, we observe a slight discrepancy between our evaluation and those reported in \citet{Zhai2020MACER}. We have verified that this is due to a sampling bias: we found \cite{Zhai2020MACER} used 500 contiguous subsamples by default in the official code, while our evaluation uses the full CIFAR-10 test set.}}
For the SmoothAdv models, we report the results for two different configurations: (a) for a fixed, pre-defined configuration across $\sigma$, and (b) for the ``best'' configuration per each $\sigma$, which is heavily examined by \citet{nips_salman19} over hundreds of models. In case of $\sigma=0.25$, however, we only report (b) as they show nearly identical results. For (a), we consider a 10-step PGD attack constrained in $\ell_2$ ball of radius $\varepsilon=1.0$, using $m=8$ noise samples.\footnote{The detailed configurations for (b), the best models, are specified in the supplementary material.}
{In case of stability training \cite{li2019stab}, we report the best models in terms of ACR across varying $\lambda$ tested: namely, we consider $\lambda\in\{1, 2, 5, 10, 20\}$ for each $\sigma$, and report $\lambda=2$ for $\sigma=0.25, 0.5$ and $\lambda=1$ for $\sigma=1.0$. The full results can be found in the supplementary material.}

{The results are presented in Table~\ref{tab:cifar10}. We also plot certified accuracy over the full range of radii per $\sigma$ in Figure~\ref{fig:cifar10}. Overall, we observe that our consistency regularization significantly and consistently improves Gaussian and SmoothAdv baselines, both in certified test accuracy and ACR.} Specifically, when $\sigma=0.50$, we found our regularization with $\lambda=10$ applied on the na\"ive Gaussian baseline could surpass the best-performing SmoothAdv model reported in ``SmoothAdv + Hyperparameter search'', in terms of ACR. Furthermore, in case of $\sigma=1.00$, consistency regularization upon the best SmoothAdv model even further improve the current state-of-the-art baseline by a significant margin, which verifies an orthogonal 
contribution of our method compared to the prior work. These observations suggest our method works better on more complex tasks, where forcing ``confident'' prediction (as done in the prior works) might be difficult. 
{We also notice that, despite its similarity with our method, the stability training \cite{li2019stab} itself does not improve ACRs even compared to the Gaussian baselines. This is because this training \eqref{eq:stab} would require $f$ to perform well both in $x$ and $x+\delta$, which is harder to force compared to that of \eqref{eq:overall_detailed} in the context of randomized smoothing.}

\vspace{\ssNMargin}
\vspace{-0.03in}
\subsection{Results on ImageNet}
\label{ss:imagenet}
\vspace{-0.03in}
\vspace{\ssNMargin}

We also evaluate our regularization scheme on ImageNet classification dataset, to show that our method is scalable on large-scale datasets. We train each model on $\sigma\in\{0.5, 1.0\}$ for 90 epochs.
We perform our evaluation on a subsampled test dataset of 500 samples as done by \citet{pmlr-v97-cohen19c}. As presented in Table~\ref{tab:imagenet}, we observe that consistency regularization still effectively improves the certified robustness, both in terms of ACR and certified test accuracy, despite its simple and efficient nature of our method. 
{Compared to the best results of SmoothAdv \cite{nips_salman19}, our results achieve a comparable robustness, despite using a single fixed configuration of hyperparameter, namely $\lambda=5$.}

\vspace{\ssNMargin}
\vspace{-0.03in}
\subsection{Runtime analysis}
\label{ss:runtime}
\vspace{-0.03in}
\vspace{\ssNMargin}

\begin{wraptable}[10]{r}{0.53\linewidth}
\centering
\vspace{-0.27in}
\caption{Comparison of training time statistics on CIFAR-10 with $\sigma=0.50$. All the baselines are trained on their official implementations separately.}
\label{tab:training}
\vspace{0.04in}
\small
    \begin{tabular}{lccccc}
        \toprule
        Models & \# HP & ACR & Mem. & Time (h) \\
        \midrule
        Gaussian &  & 0.525 &  2.9G & 4.6 \\
        \textbf{+ Consistency}  & \textbf{2} & \textbf{0.720} & 2.9G & \textbf{8.7} \\
        \midrule
        SmoothAdv  & 4 & 0.717 & 3.0G & 23.1  \\ 
        MACER  & 4 & 0.691 & 9.4G & 14.1 \\ 
        \bottomrule
    \end{tabular}
\end{wraptable}
With much effectiveness on the certified robustness, consistency regularization also offers a great efficiency in terms of training costs compared to other competitive methods. 
We compare our method with the baselines in several training statistics, including the number of hyper-parameter (\# HP), ACR, memory usage in GPU on peak computation (Mem.), and the total training time (Time). 
In this experiment, every model is trained on CIFAR-10 using one GPU of NVIDIA TITAN X (Pascal). We use $\sigma=0.5$ with hyperparameters specified in Section~\ref{ss:cifar10}. In case of SmoothAdv, we choose the best-performing configuration for training. 
Our method to compare runs upon the Gaussian baseline with $m=2$ and $\lambda=10$.

The results in Table~\ref{tab:training} show that our regularization indeed costs about twice the Gaussian baseline due to additional sampling, but one can immediately notice that this overhead is far less than others, e.g., compared to adversarial training. Furthermore, our method even achieves better ACR than other methods, which verifies a clear efficiency of consistency regularization compared to the prior work. 

\vspace{\ssNMargin}
\vspace{-0.03in}
\subsection{Ablation study}
\label{ss:ablation}
\vspace{-0.03in}
\vspace{\ssNMargin}

We conduct an ablation study for a detailed analysis on our method. 
Unless otherwised noted, we perform experiments on MNIST in this section. When consistency regularization is used, we assume it is applied upon Gaussian training. We report all the detailed results in the supplementary material.

\begin{figure}[t]
	\centering
	\subfigure[Design choices on loss]
	{
	    \includegraphics[width=0.31\linewidth]{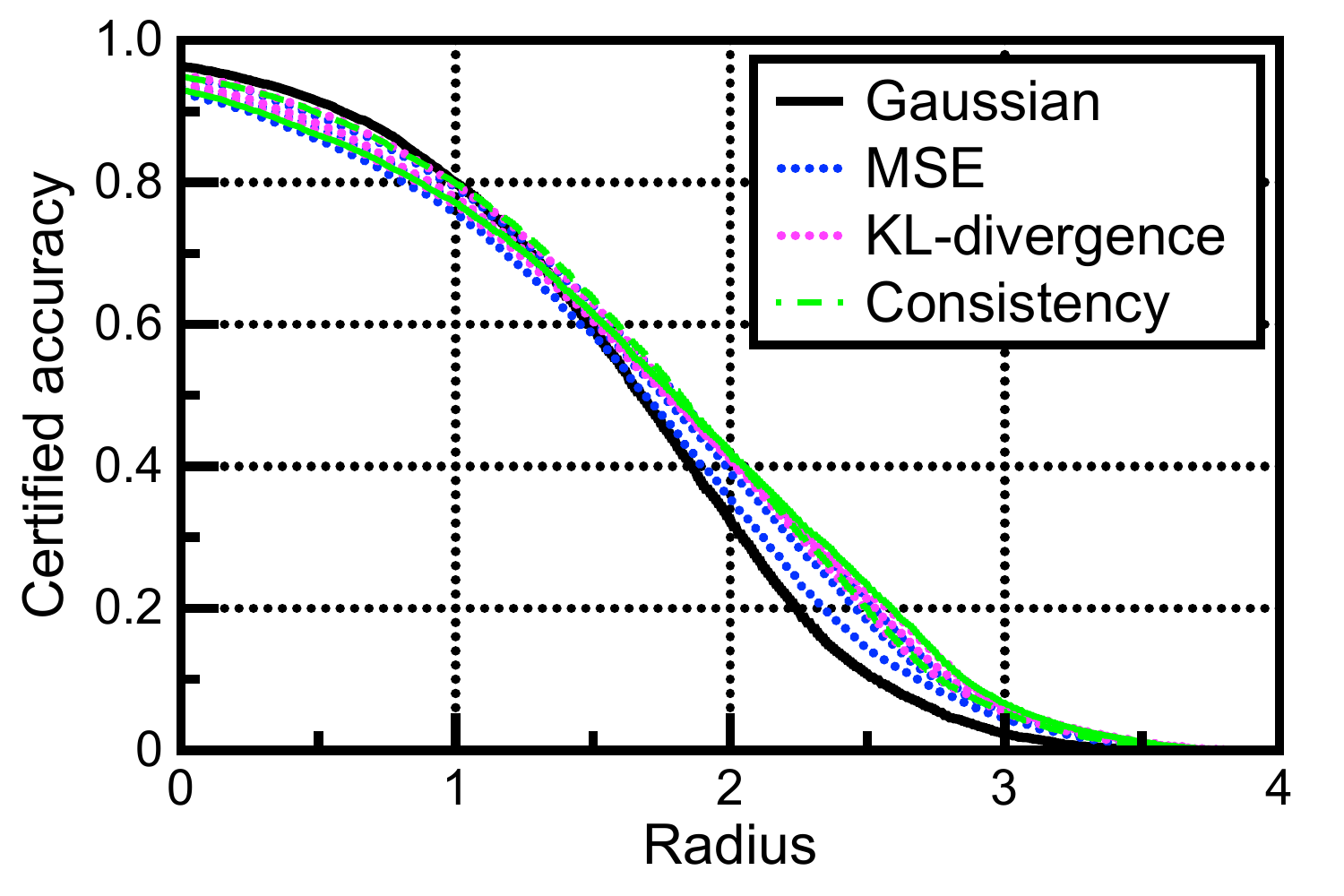}
		\label{fig:ab/loss_type}
	}
	\subfigure[Effect of $m$]
	{
	    \includegraphics[width=0.31\linewidth]{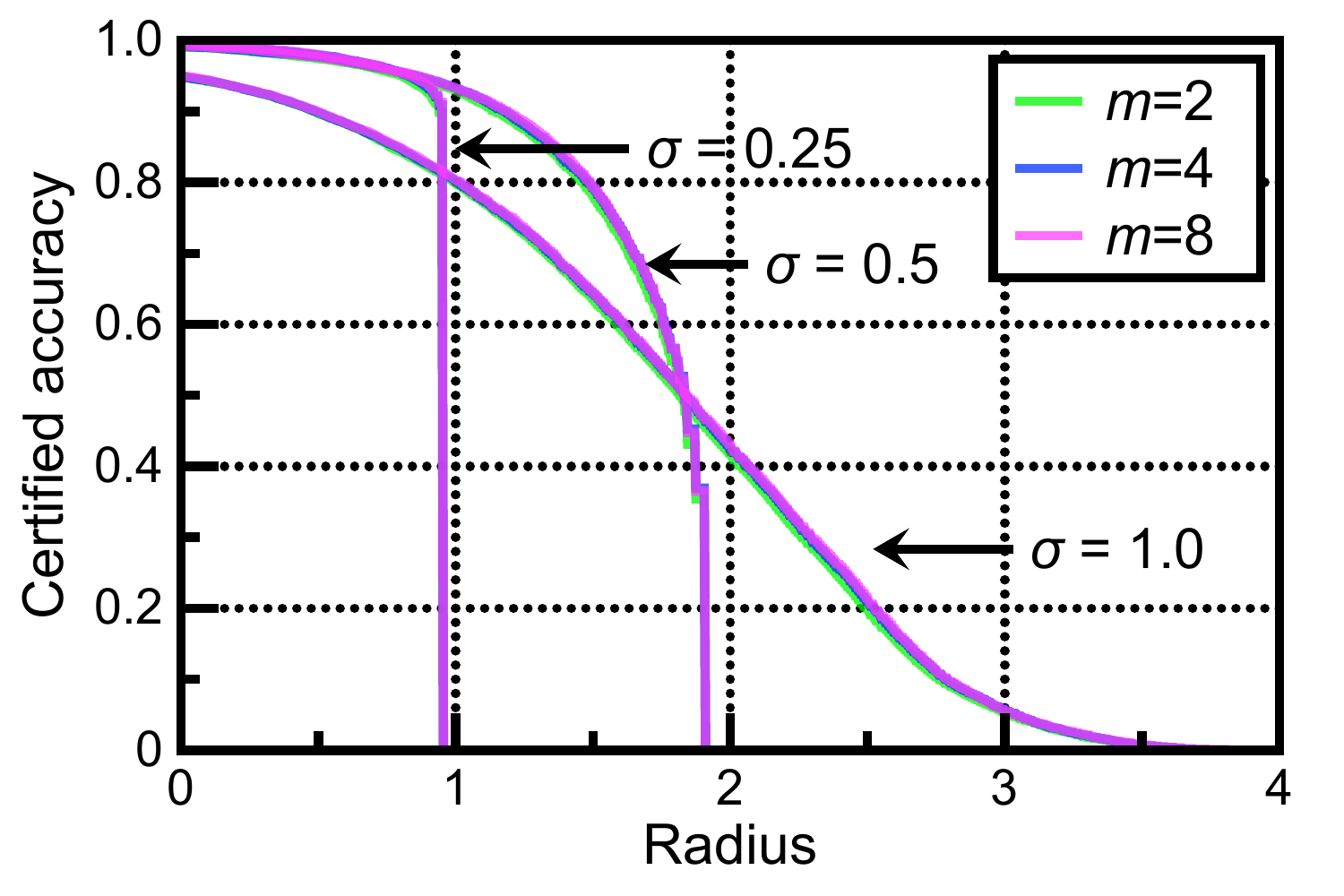}
		\label{fig:ab/effect_m}
	}
	\subfigure[Effect of $\lambda$]
	{
	    \includegraphics[width=0.31\linewidth]{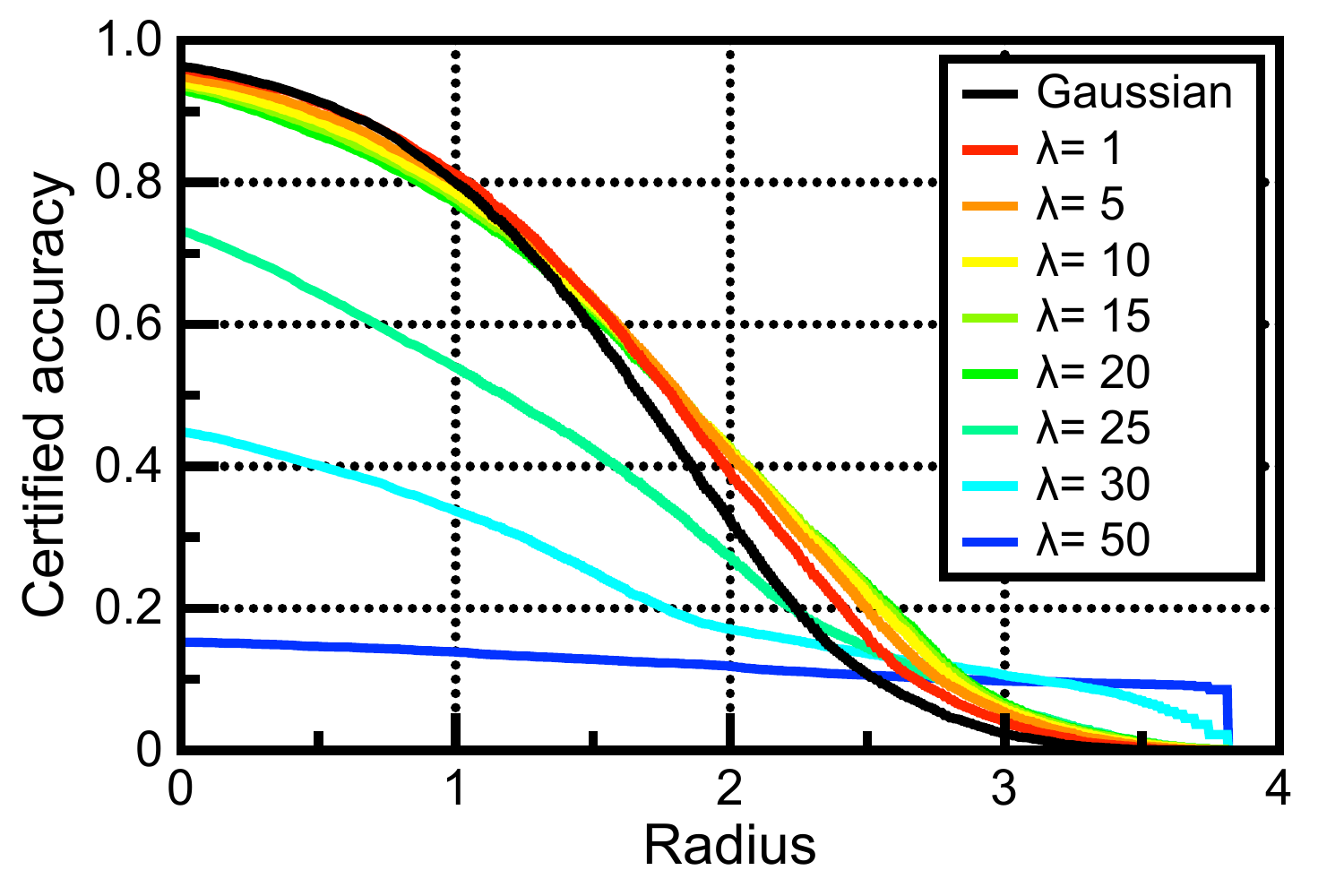}
		\label{fig:ab/effect_lbd}
	}
	\caption{Comparison of approximate certified accuracy via randomized smoothing across various types of ablation models. A sharp drop of certified accuracy in the plots exists since there is a hard upper bound that \textsc{Certify} can output for a given $\sigma$ and $n=100,000$.}
	\label{fig:ablation}
	\vspace{-0.2in}
\end{figure}

\begin{table}[t]
\centering
{\caption{Comparison of ACR on CIFAR-10 with $\sigma=0.5$ for different architectures. Bold indicates the best ACR value per architecture. For ``+ Best HP (ResNet-110)'', we use the hyperparameters those are optimized for ResNet-110 by \citet{nips_salman19}, as used in Table~\ref{tab:cifar10}.}}
\label{tab:cifar10_arch}
    \vspace{-0.02in}
    \begin{adjustbox}{width=1\linewidth}
    \begin{tabular}{clccccccccc}
    \toprule
    Architecture &  CIFAR-10 ($\sigma=0.5$) & ACR & 0.00 & 0.25 & 0.50 & 0.75 & 1.00 & 1.25 & 1.50 & 1.75 \\ 
    \midrule
    \multirow{7}{*}{ResNet-20}& Gaussian \cite{pmlr-v97-cohen19c} & 0.524 & 67.0 & 55.4 & 42.8 & 31.4 & 22.0 & 13.9 & 8.1 & 3.8  \\
    & \textbf{+ Consistency ($\lambda=10$)} & \textbf{0.686} & 60.5 & 54.4 & \textbf{47.7} & \textbf{40.9} & \textbf{34.4} & \textbf{28.0} & \textbf{22.2} & \textbf{16.5} \\
    \cmidrule(l){2-2} \cmidrule(l){3-3} \cmidrule(l){4-11}
    & SmoothAdv \cite{nips_salman19} & \underline{0.692} & 63.0 & 56.5 & 48.9 & 41.8 & 34.9 & 28.1 & 21.3 & 14.6  \\
    & + Best HP (ResNet-110)  & 0.682 & 50.8 & 47.0 & 43.1 & 39.0 & 35.1 & 31.6 & 27.2 & 22.7 \\ 
    \cmidrule(l){2-2} \cmidrule(l){3-3} \cmidrule(l){4-11}
    & Stability training \cite{li2019stab} & 0.499 & 60.2 & 50.1 & 40.5 & 31.0 & 22.1 & 14.9 & 8.2 & 3.7  \\ 
    & MACER  \cite{Zhai2020MACER} & 0.661 & 63.0 & 55.7 & 48.2 & 40.5 & 32.6 & 25.5 & 18.7 & 11.9 \\  
    \midrule
    \multirow{7}{*}{DenseNet-40}& Gaussian \cite{pmlr-v97-cohen19c} & 0.494 & 65.0 & 53.7 & 41.2 & 29.6 & 19.6 & 12.4 & 6.9 & 3.1 \\
    & \textbf{+ Consistency ($\lambda=10$)} & \textbf{0.661} & 59.1 & 52.7 & \textbf{46.1} & \textbf{39.3} & \textbf{32.8} & \textbf{27.0} & \textbf{21.1} & \textbf{15.6} \\
    \cmidrule(l){2-2} \cmidrule(l){3-3} \cmidrule(l){4-11}
    & SmoothAdv \cite{nips_salman19} & \underline{0.671} & 61.6 & 55.3 & 48.0 & 40.3 & 33.2 & 26.4 & 20.4 & 14.3 \\
    & + Best HP (ResNet-110)  & 0.659 & 49.4 & 45.8 & 41.5 & 37.6 & 33.9 & 30.2 & 26.5 & 22.1 \\ 
    \cmidrule(l){2-2} \cmidrule(l){3-3} \cmidrule(l){4-11}
    & Stability training \cite{li2019stab} & 0.497 & 56.5 & 47.9 & 38.8 & 30.7 & 23.0 & 16.5 & 9.9 & 4.9  \\
    & MACER  \cite{Zhai2020MACER} & 0.641 & 62.0 & 54.5 & 46.7 & 39.1 & 31.8 & 24.8 & 17.8 & 11.4  \\ 
    \bottomrule
    \end{tabular}
    \end{adjustbox}
    \vspace{-0.15in}
\end{table}

\textbf{Design choices on loss. }
We first examine two other popular designs for consistency regularization instead of \eqref{eq:reg}, namely, mean-squared-error \cite{sajjadi2016regularization} and KL-divergence \cite{miyato2018virtual} as follow:
\begin{equation}
    L^{\tt MSE} := \lambda\cdot||F(x+\delta_1) - F(x+\delta_2) ||_2^2 \quad \text{ and} \quad
    L^{\tt KL} := \lambda\cdot\mathbb{E}_\delta[\mathrm{KL}(\hat{F}(x)\ ||\ F(x + \delta))],
\end{equation}
where $\delta_1, \delta_2 \sim \mathcal{N}(0, \sigma^2 I)$. We evaluate certified test accuracy on $\sigma=1.00$ for these regularization with varying $\lambda\in\{5, 20, 50\}$, and compare the results with $L^{\tt con}$ with $\lambda\in\{5, 20\}$. The results are presented in Figure~\ref{fig:ab/loss_type}. In general, we observe that both regularizers, namely $L^{\tt MSE}$ and $L^{\tt KL}$, are also capable to improve the certified robustness, but they could not achieve a better ACR than $L^{\tt con}$ even with a moderately large $\lambda$. Considering that $L^{\tt KL}$ is equivalent to $L^{\tt con}$ when $\eta=0$ in \eqref{eq:reg}, this observation indicates the importance of regularizing the \emph{entropy} of the mean prediction. Indeed, we empirically observe that both $L^{\tt MSE}$ and $L^{\tt KL}$ often lead the predictions to be too close to the uniform when $\lambda$ is large, which may harm the discriminative performance of the base classifier.

\textbf{Effect of $m$. }
As mentioned in Section~\ref{s:method}, the computational costs for our regularization scheme highly depends on the number of noise samples used, namely $m>1$. Nevertheless, we observe that our regularization is fairly robust on the choice of $m$, so that $m=2$ usually leads to good enough performance. In Figure~\ref{fig:ab/effect_m}, we compare the certified robustness of models trained with our regularization with varying $m\in\{2, 4, 8\}$. For each $m$, we present three different models under various $\sigma\in\{0.25, 0.5, 1.0\}$. The results show that models using $m=2$ perform nearly identically to others, while one could observe slight improvements for larger $m$.
In practice, this observation reduces much of the hyperparameter complexity in our method: by simply letting $m$ to be small, e.g., $m=2$, while fixing $\eta=0.5$, $\lambda$ becomes the only crucial hyperparameter.

\textbf{Effect of $\lambda$. } 
We also investigate the effect of having different $\lambda$ in Figure~\ref{fig:ab/effect_lbd}. As expected, we observe a clear trade-off between accuracy and robustness of the corresponding smoothed classifier by controlling $\lambda$. Furthermore, for a sufficiently large $\lambda$, e.g., $\lambda=50$ in Figure~\ref{fig:ab/effect_lbd}, a classifier is often trained to return maximal certifiable radius for any input when smoothed, even the accuracy falls into chance-level. Nevertheless, this would be a desirable property for a trade-off term between accuracy and robustness, which has not been explored much for smoothed classifiers. 

{\textbf{Different architectures. } In our experiments, we follow the prior works \cite{pmlr-v97-cohen19c, nips_salman19, Zhai2020MACER} to choose network architectures for a fair comparison, e.g., we use ResNet-110 \cite{he2016deep} for CIFAR-10. To explore the effect of different architectures, we further test our method with ResNet-20 \cite{he2016deep} and \mbox{DenseNet-40} \cite{huang2017densely} on CIFAR-10, as summarized in Table~\ref{tab:cifar10_arch}. We assume $\sigma=0.5$ for this experiment, and use the same hyperparameters specified in Section~\ref{ss:cifar10}. Overall, our method also consistently outperforms other baselines except SmoothAdv on these architectures. In particular, it is remarkable that our choice of hyperparameter, namely $\lambda=10$, transfers well to other architectures, compared to SmoothAdv: the best working hyperparameters for ResNet-110 could not further improve SmoothAdv from the baseline configuration, i.e., it may require a further optimization to perform better.}

\vspace{\sNMargin}
\vspace{-0.05in}
\section{Conclusion}
\vspace{-0.05in}
\vspace{\sNMargin}

In this paper, we show \emph{consistency regularization} can play a key role in certifiable robustness of smoothed classifiers. We think our work would emphasize the importance of \emph{noise-consistent} inference in deep neural networks, one of under-explored topics despite its desirable property. We also expect our work can be a useful guideline when other researchers will study the noise-consistency in other problems in the future. Many questions are related: how can we design a noise-invariant neural network, or for which family of noise this would be allowed, just to name a few.

\section*{Broader Impact}

The potential risk of adversarial attacks has left many practitioners hesitant to apply the latest developments in deep learning into their systems. 
Adversarial robustness of deep neural networks is one of the most important research problems toward \emph{AI safety} \cite{amodei2016concrete}, with much impact on various applications especially for security-concerned systems: e.g., medical diagnosis \cite{caruana2015intelligible}, speech recognition \cite{qin2019speech}, and autonomous driving \cite{yurtsever2020survey}.
{Our research could be beneficial for those who design such systems, thanks to the certifiable guarantees on adversarial robustness that \emph{randomized smoothing} can provide.
Especially, we expect the simplicity of our method would encourage many practitioners to incorporate randomized smoothing into their systems along with our work. A practical success of systems equipped with a sufficient amount of certified robustness would be fatal for those who maliciously attempt to break down the system via adversarial attacks.}

This statement, however, presumes that many of the practical issues on the current randomized smoothing technique would be resolved in future research. 
For example, (a) randomized smoothing requires exponentially many inferences for a single reliable inference, and (b) there is still a gap between theoretical guarantee \cite{dvijotham2020a, yang2020randomized}  and practice \cite{xiao2018spatially, bhattad2020Unrestricted} on robustness that randomized smoothing currently gives: 
consequently, current randomized smoothing can be easily misused in practical systems, 
{and a failure of such systems may implicitly lead practitioners to have a biased, false sense of security.}
We believe our research is a step toward reducing this practical gap to deploy randomized smoothing into the real-world.

\begin{ack}
This work was conducted by Center for Applied Research in Artificial Intelligence(CARAI) grant funded by Defense Acquisition Program Administration(DAPA) and Agency for Defense Development(ADD) (UD190031RD).
\end{ack}

\small

\bibliographystyle{plainnat}
\bibliography{references}

\end{document}


\onecolumn
    \clearpage
    \begin{center}
    {\bf {\LARGE Supplementary Material:}} \\
    \vspace{0.15in}
    {\bf {\Large Consistency Regularization for Certified Robustness of Smoothed Classifiers}}
    \end{center}
    
    \appendix
    
    \section{Details on experimental setups}
    
    \subsection{Training details}
    
    We train every model via stochastic gradient descent (SGD) with Nesterov momentum of weight 0.9 without dampening. We set a weight decay of $10^{-4}$ for all the models. We use different training schedules for each dataset: (a) MNIST: The initial learning rate is set to 0.01; We train a model for 90 epochs with mini-batch size 256, and the learning rate is decayed by 0.1 at 30-th and 60-th epoch, (b) CIFAR-10: The initial learning rate is set to 0.1; We train a model for 150 epochs with mini-batch size 256, and the learning rate is decayed by 0.1 at 50-th and 100-th epoch, and (c) ImageNet: The initial learning rate is set to 0.1; We train a model for 90 epochs with mini-batch size 200, and the learning rate is decayed by 0.1 at 30-th and 60-th epoch. 
    When SmoothAdv is used, we adopt the \emph{warm-up} strategy on attack radius $\varepsilon$  \cite{nips_salman19}, i.e., $\varepsilon$ is initially set to zero, and linearly increased during the first 10 epochs to a pre-defined hyperparameter.
    
    \subsection{Datasets}
    
    \textbf{MNIST} dataset \citep{dataset/mnist} consists 70,000 gray-scale hand-written digit images of size 28$\times$28, 60,000 for training and 10,000 for testing. Each of the images is labeled from 0 to 9, i.e., there are 10 classes. 
    When training on MNIST, we do not perform any pre-processing except for normalizing the range of each pixel from 0-255 to 0-1. 
    The full dataset can be downloaded at \url{http://yann.lecun.com/exdb/mnist/}.
    
    \textbf{CIFAR-10} dataset \citep{dataset/cifar} consist of 60,000 RGB images of size 32$\times$32 pixels, 50,000 for training and 10,000 for testing. Each of the images is labeled to one of 10 classes, 
    and the number of data per class is set evenly, i.e., 6,000 images per each class. 
    We follow the same data-augmentation scheme used in \citet{pmlr-v97-cohen19c, nips_salman19} for a fair comparison, namely, we use random horizontal flip and random translation up to 4 pixels. 
    We also normalize the images in pixel-wise by the mean and the standard deviation calculated from the training set. Here, an important practical point is that this normalization is done \emph{after} a noise is added to input when regarding randomized smoothing, following \citet{pmlr-v97-cohen19c}. This is to ensure that noise is given to the original image coordinates. In practical implementations, this can be done by placing the normalization as the first layer of base classifiers, instead of as a pre-processing step.
    The full dataset can be downloaded at \url{https://www.cs.toronto.edu/~kriz/cifar.html}
    
    \textbf{ImageNet} classification dataset \cite{ILSVRC15} consists of 1.2 million training images and 50,000 validation images, which are labeled by one of 1,000 classes. 
    For data-augmentation, we perform 224$\times$224 random cropping with random resizing and horizontal flipping to the training images. At test time, on the other hand, 224$\times$224 center cropping is performed after re-scaling the images into 256$\times$256. This pre-processing scheme is also used in \citet{pmlr-v97-cohen19c, nips_salman19} as well.
    Similar to CIFAR-10, all the images are normalized \emph{after} adding a noise in pixel-wise by the pre-computed mean and standard deviation.
    A link for downloading the full dataset can be found in \url{http://image-net.org/download}.
    
    \begin{table*}[ht]
\centering
\caption{Detailed specification of hyperparameters used in the best-performing SmoothAdv models.}
\label{tab:smoothadv}
\vspace{0.03in}
    \begin{tabular}{cccccc}
    \toprule
    Dataset & $\sigma$ & Method & \# steps & $\varepsilon$ & $m$  \\ 
    \midrule
    \multirow{3}{*}{CIFAR-10}& 0.25 & PGD & 10 & 255 & 4  \\
    & 0.50 & PGD & 10 & 512 & 2  \\
    & 1.00 & PGD  & 10 & 512 & 2  \\  
    \cmidrule{1-2}  \cmidrule(l){3-6}
    \multirow{2}{*}{ImageNet}& 0.50 & PGD & 1 & 255 & 1 \\
    & 1.00 & PGD & 1 & 512 & 1 \\
    \bottomrule
    \end{tabular}
\end{table*}
    
    \subsection{Detailed configurations of SmoothAdv models}
    
    In Table~\ref{tab:smoothadv}, we specify the exact configurations used in our evaluation for the best-performing SmoothAdv models. These configurations have originally explored by \citet{nips_salman19}
    via a grid search over 4 hyperparameters: namely, (a) attack method (Method): PGD \cite{madry2018towards} or DDN \cite{Rony_2019_CVPR}, (b) the number of steps (\# steps), (c) the maximum allowed $\ell_2$ perturbation on the input ($\varepsilon$), and (d) the number of noise samples ($m$). We choose one pre-trained model per $\sigma$ for CIFAR-10 and ImageNet, among those officially released and classified as the best-performing models by \citet{nips_salman19}. The link to download all the pre-trained models can be found in \url{https://github.com/Hadisalman/smoothing-adversarial}. 
    
    \begin{table}[t]
\centering
\caption{Comparison of approximate certified test accuracy on MNIST dataset. For each model, training and certification are done with the same smoothing factor specified in $\sigma$. Each of the values indicates the fraction of test samples those have $\ell_2$ certified radius larger than the threshold specified at the top row. We set our result bold-faced whenever the value improves the baseline. For ACR, we underlined the best-performing model per each $\sigma$.}
\label{tab:mnist}
    \begin{adjustbox}{width=1\linewidth}
    \begin{tabular}{clcccccccccccc}
    \toprule
    $\sigma$ &  Models (MNIST) & ACR & 0.00 & 0.25 & 0.50 & 0.75 & 1.00 & 1.25 & 1.50 & 1.75 & 2.00 & 2.25 & 2.50 \\ 
    \midrule
    \multirow{7}{*}{0.25}& Gaussian \cite{pmlr-v97-cohen19c} & 0.911 & 99.2 & 98.5 & 96.7 & 93.3 & 0.0 & 0.0 & 0.0 & 0.0 & 0.0 & 0.0 & 0.0 \\
    & \textbf{+ Consistency ($\lambda=10$)}  & \textbf{0.928} & \textbf{99.5} & \textbf{98.9} & \textbf{98.0} & \textbf{96.0} & 0.0 & 0.0 & 0.0 & 0.0 & 0.0 & 0.0 & 0.0 \\
    \cmidrule(l){2-2} \cmidrule(l){3-3} \cmidrule(l){4-14}
    & SmoothAdv \cite{nips_salman19} & {\underline{0.932}} & 99.4 & 99.0 & 98.2 & 96.8 & 0.0 & 0.0 & 0.0 & 0.0 & 0.0 & 0.0 & 0.0  \\
    & \textbf{+ Consistency ($\lambda=1$)} & 0.932 & 99.3 & 98.9 & 98.1 & 96.8 & 0.0 & 0.0 & 0.0 & 0.0 & 0.0 & 0.0 & 0.0  \\
    \cmidrule(l){2-2} \cmidrule(l){3-3} \cmidrule(l){4-14}
    & Stability training  \cite{li2019stab} & 0.915 & 99.3 & 98.6 & 97.1 & 93.8 & 0.0 & 0.0 & 0.0 & 0.0 & 0.0 & 0.0 & 0.0 \\
    & MACER  \cite{Zhai2020MACER} & 0.920 & 99.3 & 98.7 & 97.5 & 94.8 & 0.0 & 0.0 & 0.0 & 0.0 & 0.0 & 0.0 & 0.0  \\ 
    \midrule
    \multirow{7}{*}{0.50}& Gaussian \cite{pmlr-v97-cohen19c} & 1.553 & 99.2 & 98.3 & 96.8 & 94.3 & 89.7 & 81.9 & 67.3 & 43.6 & 0.0 & 0.0 & 0.0  \\
    & \textbf{+ Consistency ($\lambda=5$)}  & \textbf{1.657} & 99.2 & \textbf{98.6} & \textbf{97.6} & \textbf{95.9} & \textbf{93.0} & \textbf{87.8} & \textbf{78.5} & \textbf{60.5} & 0.0 & 0.0 & 0.0  \\
    \cmidrule(l){2-2} \cmidrule(l){3-3} \cmidrule(l){4-14}
    & SmoothAdv \cite{nips_salman19} & 1.687 & 99.0 & 98.3 & 97.3 & 95.8 & 93.2 & 88.5 & 81.1 & 67.5 & 0.0 & 0.0 & 0.0  \\
    & \textbf{+ Consistency ($\lambda=1$)} & \textbf{\underline{1.697}} & 98.6 & 98.1 & 97.0 & 95.3 & 92.7 & 88.5 & \textbf{82.2} & \textbf{70.5} & 0.0 & 0.0 & 0.0  \\
    \cmidrule(l){2-2} \cmidrule(l){3-3} \cmidrule(l){4-14}
    & Stability training  \cite{li2019stab} & 1.570 & 99.2 & 98.5 & 97.1 & 94.8 & 90.7 & 83.2 & 69.2 & 45.4 & 0.0 & 0.0 & 0.0 \\
    & MACER  \cite{Zhai2020MACER} & 1.594 & 98.5 & 97.5 & 96.2 & 93.7 & 90.0 & 83.7 & 72.2 & 54.0 & 0.0 & 0.0 & 0.0  \\
    \midrule
    \multirow{7}{*}{1.00}& Gaussian \cite{pmlr-v97-cohen19c} & 1.620 & 96.4 & 94.4 & 91.4 & 87.0 & 79.9 & 71.0 & 59.6 & 46.2 & 32.6 & 19.7 & 10.8  \\
    & \textbf{+ Consistency ($\lambda=5$)} & \textbf{1.740} & 95.0 & 93.0 & 89.7 & 85.4 & 79.7 & \textbf{72.7} & \textbf{63.6} & \textbf{53.0} & \textbf{41.7} & \textbf{30.8} & \textbf{20.3}  \\
    \cmidrule(l){2-2} \cmidrule(l){3-3} \cmidrule(l){4-14}
    & SmoothAdv \cite{nips_salman19} & 1.779 & 95.8 & 93.9 & 90.6 & 86.5 & 80.8 & 73.7 & 64.6 & 53.9 & 43.3 & 32.8 & 22.2  \\
    & \textbf{+ Consistency ($\lambda=1$)} & \textbf{\underline{1.819}} & 94.2 & 92.0 & 88.6 & 84.3 & 79.0 & 72.1 & 64.0 & \textbf{54.6} & \textbf{45.5} & \textbf{37.2} & \textbf{28.0}  \\
    \cmidrule(l){2-2} \cmidrule(l){3-3} \cmidrule(l){4-14}
    & Stability training  \cite{li2019stab} & 1.634 & 96.5 & 94.6 & 91.7 & 87.4 & 80.6 & 72.0 & 60.5 & 46.8 & 33.1 & 20.0 & 11.2 \\
    & MACER \cite{Zhai2020MACER} & 1.570 & 92.0 & 88.5 & 84.0 & 78.1 & 71.5 & 63.8 & 55.3 & 46.3 & 36.5 & 26.2 & 16.3  \\ 
    \bottomrule
    \end{tabular}
    \end{adjustbox}
\end{table}
    \begin{figure}[t]
	\centering
	\subfigure[$\sigma=0.25$]
	{
	    \includegraphics[width=0.31\linewidth]{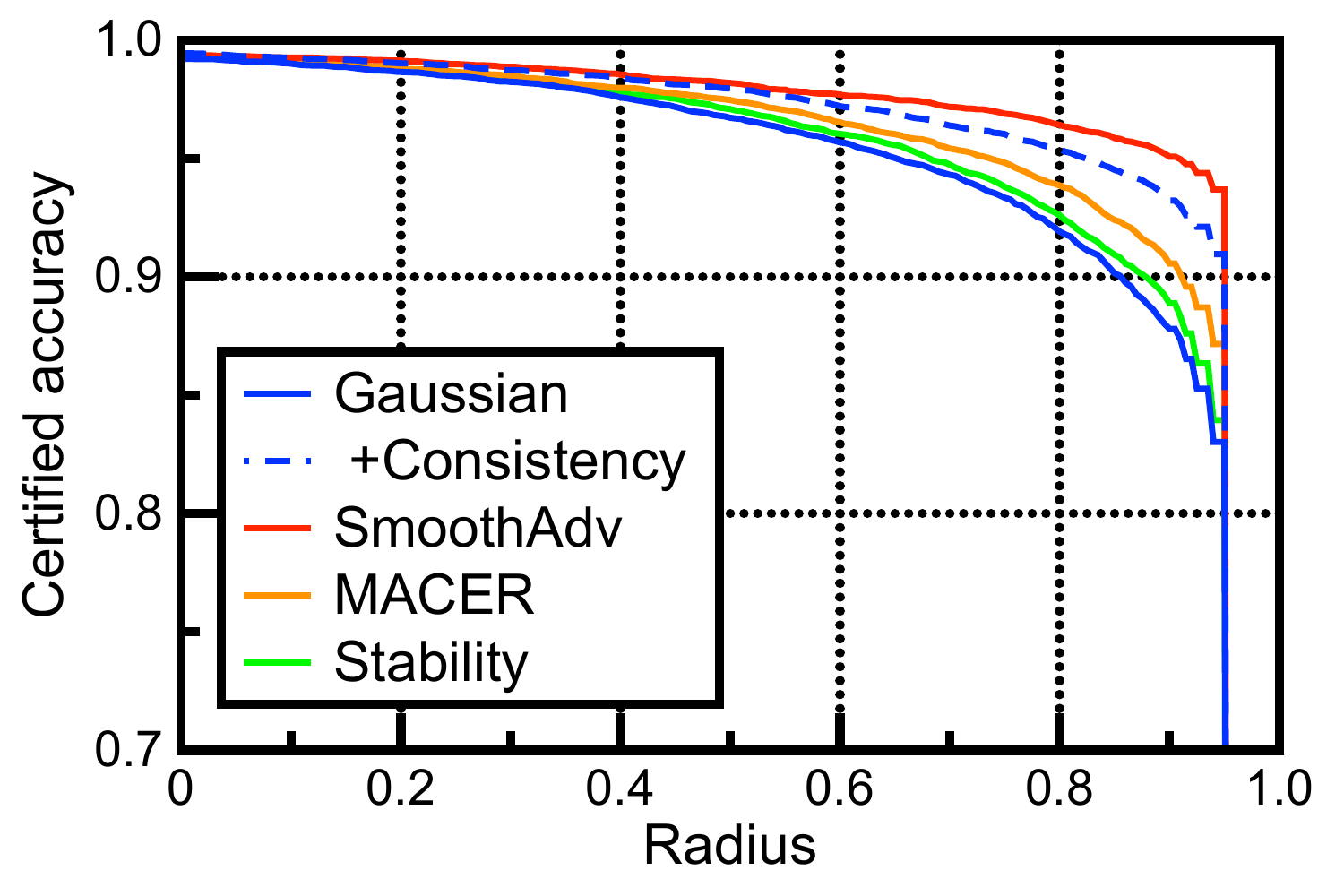}
		\label{fig:mnist_25}
	}
	\subfigure[$\sigma=0.50$]
	{
	    \includegraphics[width=0.31\linewidth]{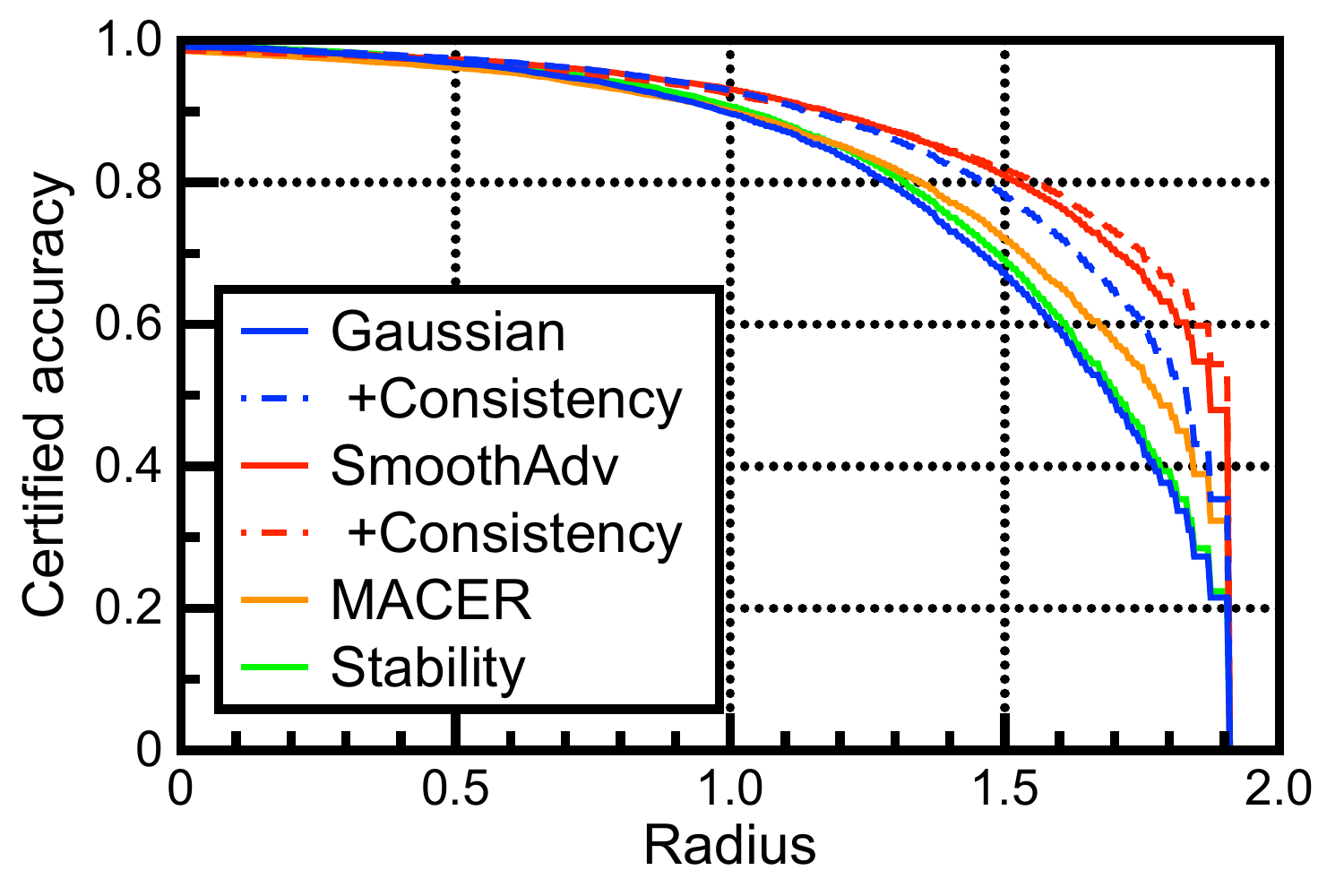}
		\label{fig:mnist_50}
	}
	\subfigure[$\sigma=1.00$]
	{
	    \includegraphics[width=0.31\linewidth]{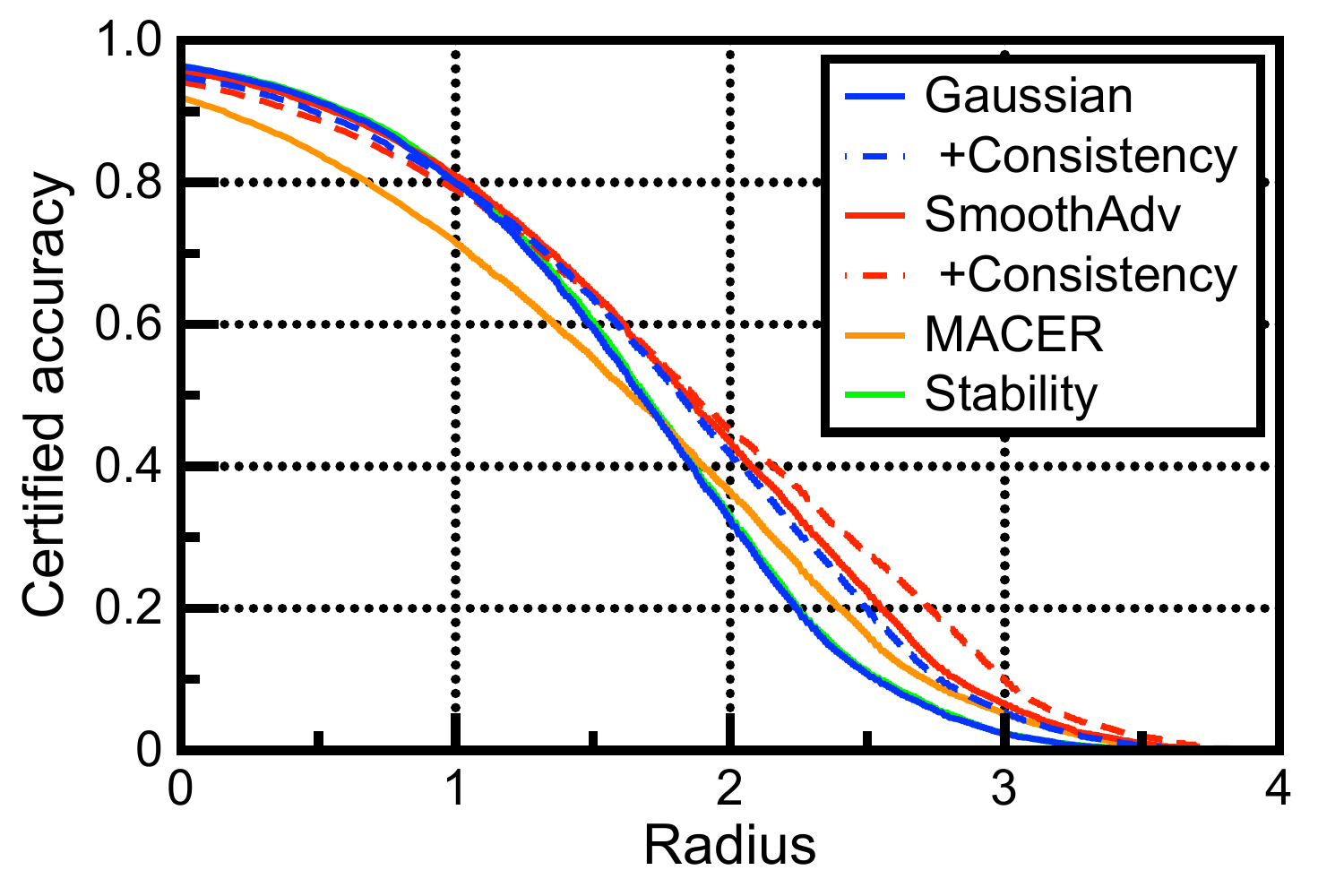}
		\label{fig:mnist_100}
	}
	\caption{Comparison of approximate certified accuracy via randomized smoothing for various training methods on MNIST. A sharp drop of certified accuracy in the plots exists since there is a hard upper bound that \textsc{Certify} can output for a given $\sigma$ and $n=100,000$.}
	\label{fig:mnist}
    \end{figure}
    
    \section{Results on MNIST}
    
    We train every MNIST model for 90 epochs. 
    We consider a fixed configuration of hyperparameters when SmoothAdv is used in MNIST: specifically, we perform a 10-step projected gradient descent (PGD) attack constrained in $\ell_2$ ball of radius $\varepsilon=1.0$ for each input, while the objective is approximated with $m=4$ noise samples. 
    For the MACER models, on the other hand, we generally follow the hyperparameters specified in the original paper \cite{Zhai2020MACER}: we set $m=16$, $\lambda=16.0$, $\gamma=8.0$ and $\beta=16.0$.\footnote{We refer the readers to \citet{Zhai2020MACER} for the details on each hyperparemeter.} In $\sigma=1.0$, however, we had to reduce $\lambda$ to 6 for a successful training. Nevertheless, we have verified that the ACRs computed from the reproduced models are comparable to those reported in the original paper. We use $\lambda=2$ when stability training \cite{li2019stab} is applied in this section.
    
    We report the results in Table~\ref{tab:mnist} and Figure~\ref{fig:mnist}. Overall, we observe that our consistency regularization stably improve Gaussian and SmoothAdv baselines in ACR, except when applied to SmoothAdv on $\sigma=0.25$. This corner-case is possibly due to that the model is already achieve to the best capacity via SmoothAdv, regarding that MNIST on $\sigma=0.25$ is relatively a trivial task. For the rest non-trivial cases, nevertheless, our regularization shows a remarkable effectiveness in two aspects: (a) applying our consistency regularization on Gaussian, the simplest baseline, dramatically improves the certified test accuracy and ACR even outperforming the recently proposed MACER by a large margin, and (b) when applied to SmoothAdv, our method could further improve ACR. In particular, one could observe that our regularization significantly improves the certified accuracy especially at large radii, where a classifier should attain a high value of $p^{(1)}$ \eqref{eq:cr}, i.e., a consistent prediction is required.
    
    \section{Variance of results over multiple runs}
    
     In our experiments, we compare single-run results following other baselines considered in this paper \cite{pmlr-v97-cohen19c, nips_salman19, li2019stab, Zhai2020MACER}. In Table~\ref{tab:mnist_var}, we report the mean and standard deviation of ACRs across 5 seeds for the MNIST results reported in Table~\ref{tab:mnist}. In general, we observe ACR of a given training method is fairly robust to network initialization.
    
    \begin{table}[h]
    \centering
    \small
        \caption{Comparison of ACR for various training methods on MNIST. The reported values are the mean and standard deviation across 5 seeds. We set our result bold-faced whenever the value improves the baseline, and the underlined are best-performing model per $\sigma$.}
	    \label{tab:mnist_var}
        \begin{tabular}{lccc}
        \toprule
        ACR (MNIST) & $\sigma=0.25$ & $\sigma=0.50$ & $\sigma=1.00$  \\ 
        \midrule
        Gaussian \cite{pmlr-v97-cohen19c} & 0.9108\pms{0.0003} & 1.5581\pms{0.0016} & 1.6184\pms{0.0021} \\
        \textbf{+ Consistency}  & \textbf{0.9279\pms{0.0003}}  & \textbf{1.6549\pms{0.0011}} & \textbf{1.7376\pms{0.0017}}  \\
        \cmidrule(l){1-1} \cmidrule(l){2-4}
        SmoothAdv \cite{nips_salman19} & {0.9322}\pms{0.0005} & 1.6872\pms{0.0007} & 1.7786\pms{0.0017}  \\
        \textbf{+ Consistency} & \textbf{\underline{0.9323}}\pms{0.0001} & \textbf{\underline{1.6957}\pms{0.0005}} & \textbf{\underline{1.8163}\pms{0.0020}}   \\
        \cmidrule(l){1-1} \cmidrule(l){2-4}
        Stability \cite{li2019stab} & 0.9152\pms{0.0007} & 1.5719\pms{0.0028} & 1.6341\pms{0.0018}  \\ 
        MACER  \cite{Zhai2020MACER} & 0.9201\pms{0.0006} & 1.5899\pms{0.0069} & 1.5950\pms{0.0051}  \\ 
        \bottomrule
        \end{tabular}
    \end{table}
    
    \section{Detailed results in ablation study}
    
    We report the detailed results for the experiments performed in ablation study (see Section~4.6 in the main text). Table~\ref{tab:ab_loss}, \ref{tab:ab_m}, and \ref{tab:ab_lbd} are corresponded to Figure~4(a), 4(b), and 4(c) in the main text, respectively.
    
    \begin{table*}[ht]
\centering
\caption{Comparison of approximate certified test accuracy (\%) on MNIST, for varing loss functions and $\lambda$. We set our result bold-faced whenever the value improves the baseline. For ACR, we underlined the best-performing model.}
\label{tab:ab_loss}
\vspace{0.02in}
    \begin{adjustbox}{width=1\linewidth}
    \begin{tabular}{cccccccccccccc}
    \toprule
    Model & $\lambda$ & ACR & 0.00 & 0.25 & 0.50 & 0.75 & 1.00 & 1.25 & 1.50 & 1.75 & 2.00 & 2.25 & 2.50 \\ 
    \cmidrule{1-2} \cmidrule(l){3-3} \cmidrule(l){4-14}
    \multirow{1}{*}{Gaussian}& 0 & 1.620 & 96.4 & 94.4 & 91.4 & 87.0 & 79.9 & 71.0 & 59.6 & 46.2 & 32.6 & 19.7 & 10.8  \\  
    \cmidrule{1-2} \cmidrule(l){3-3} \cmidrule(l){4-14}
    \multirow{3}{*}{MSE}& 5 & \textbf{1.732} & 94.9 & 92.9 & 89.3 & 85.0 & 79.3 & \textbf{71.7} & \textbf{62.7} & \textbf{52.5} & \textbf{41.5} & \textbf{31.2} & \textbf{21.3}  \\
    & 20 & \textbf{1.677} & 93.6 & 91.0 & 87.5 & 83.0 & 77.1 & 69.9 & \textbf{60.8} & \textbf{50.3} & \textbf{39.5} & \textbf{28.6} & \textbf{18.4}  \\
    & 50 & 1.603 & 92.5 & 90.0 & 86.1 & 81.3 & 75.5 & 67.7 & 58.6 & \textbf{47.4} & \textbf{35.7} & \textbf{24.1} & \textbf{14.5}  \\
    \cmidrule{1-2}  \cmidrule(l){3-3} \cmidrule(l){4-14}
    \multirow{3}{*}{KL-divergence}& 5 & \textbf{1.729} & 95.2 & 93.0 & 89.9 & 85.4 & 79.6 & \textbf{72.4} & \textbf{62.9} & \textbf{52.2} & \textbf{41.1} & \textbf{30.3} & \textbf{19.6}  \\
    & 20 & \textbf{1.713} & 94.0 & 91.7 & 88.2 & 83.5 & 77.7 & 70.5 & \textbf{61.5} & \textbf{51.4} & \textbf{41.2} & \textbf{31.1} & \textbf{21.4} \\
    & 50 & \textbf{1.707} & 93.4 & 90.7 & 87.1 & 82.3 & 76.8 & 69.4 & \textbf{60.6} & \textbf{50.9} & \textbf{41.3} & \textbf{31.8} & \textbf{22.6}  \\
    \cmidrule{1-2}  \cmidrule(l){3-3} \cmidrule(l){4-14}
    \multirow{2}{*}{Cross-entropy}& 5 & \textbf{1.740} & 95.0 & 93.0 & 89.7 & 85.4 & 79.7 & \textbf{72.7} & \textbf{63.6} & \textbf{53.0} & \textbf{41.7} & \textbf{30.8} & \textbf{20.3} \\
    & 20 & \textbf{1.720} & 93.0 & 90.3 & 86.6 & 82.3 & 77.1 & 70.2 & \textbf{61.6} & \textbf{52.0} & \textbf{42.1} & \textbf{32.5} & \textbf{23.4} \\
    \bottomrule
    \end{tabular}
    \end{adjustbox}
\end{table*}
    \begin{table*}[ht]
\centering
\caption{Comparison of approximate certified test accuracy on MNIST for varying $m\in\{2, 4, 8\}$. For each model, training and certification are done with the same smoothing factor specified in $\sigma$.}
\label{tab:ab_m}
\vspace{0.02in}
    \begin{tabular}{cccccccccccccc}
    \toprule
    $\sigma$ &  $m$ & ACR & 0.00 & 0.25 & 0.50 & 0.75 & 1.00 & 1.25 & 1.50 & 1.75 & 2.00 & 2.25 & 2.50 \\ 
    \cmidrule{1-2} \cmidrule(l){3-3} \cmidrule(l){4-14}
    \multirow{3}{*}{0.25}& 2 & 0.926 & 99.4 & 98.9 & 97.8 & 95.6 & 0.0 & 0.0 & 0.0 & 0.0 & 0.0 & 0.0 & 0.0 \\
    & 4  & 0.928 & 99.5 & 98.9 & 97.9 & 96.1 & 0.0 & 0.0 & 0.0 & 0.0 & 0.0 & 0.0 & 0.0  \\
    & 8 & 0.929 & 99.4 & 99.0 & 98.0 & 96.1 & 0.0 & 0.0 & 0.0 & 0.0 & 0.0 & 0.0 & 0.0 \\
    \cmidrule{1-2} \cmidrule(l){3-3} \cmidrule(l){4-14}
    \multirow{3}{*}{0.50}& 2 & 1.657 & 99.2 & 98.6 & 97.6 & 95.9 & 93.0 & 87.8 & 78.5 & 60.5 & 0.0 & 0.0 & 0.0  \\
    & 4  & 1.666 & 99.2 & 98.6 & 97.7 & 96.0 & 93.3 & 88.2 & 79.4 & 62.3 & 0.0 & 0.0 & 0.0  \\
    & 8 & 1.667 & 99.2 & 98.7 & 97.6 & 95.9 & 93.3 & 88.6 & 79.5 & 62.1 & 0.0 & 0.0 & 0.0  \\
    \cmidrule{1-2} \cmidrule(l){3-3} \cmidrule(l){4-14}
    \multirow{3}{*}{1.00}& 2 & 1.740 & 95.0 & 93.0 & 89.7 & 85.4 & 79.7 & 72.7 & 63.6 & 53.0 & 41.7 & 30.8 & 20.3 \\
    & 4  & 1.756 & 94.9 & 92.9 & 89.8 & 85.6 & 80.2 & 73.3 & 64.5 & 54.0 & 42.7 & 31.9 & 21.0  \\
    & 8 & 1.762 & 95.0 & 93.1 & 90.0 & 85.8 & 80.3 & 73.7 & 64.6 & 54.2 & 43.1 & 32.2 & 21.5  \\
    \bottomrule
    \end{tabular}
\end{table*}
    \begin{table*}[ht]
\centering
\caption{Comparison of approximate certified test accuracy on MNIST for varying $\lambda$. We set our result bold-faced whenever the value improves the baseline ($\lambda = 0.0$). For ACR, we underlined the best-performing model.}
\label{tab:ab_lbd}
\vspace{0.02in}
    \begin{tabular}{rccccccccc}
    \toprule
    $\lambda$ & ACR & 0.00 & 0.50 & 1.00 & 1.50 & 2.00 & 2.50 & 3.00 & 3.50 \\ 
    \cmidrule{1-1} \cmidrule(lr){2-2} \cmidrule(l){3-10}
     0.0 & 1.619 & 96.3 & 91.4 & 79.8 & 59.4 & 32.5 & 10.9 & 2.4 & 0.0 \\
    \cmidrule{1-1} \cmidrule(lr){2-2} \cmidrule(l){3-10}
     1.0  & \textbf{1.714} & 96.0 & 91.2 & \textbf{81.1} & \textbf{63.5} & \textbf{39.2} & \textbf{16.2} & \textbf{4.2} & \textbf{0.4} \\
     5.0 & \textbf{1.740} & 95.0 & 89.7 & \textbf{79.9} & \textbf{63.7} & \textbf{41.9} & \textbf{20.0} & \textbf{5.4} & \textbf{0.6} \\
     10.0 & \textbf{1.735} & 94.1 & 88.6 & 78.5 & \textbf{62.8} & \textbf{42.4} & \textbf{22.1} & \textbf{5.9} & \textbf{0.9}  \\
     15.0  & \textbf{1.731} & 93.6 & 87.7 & 77.8 & \textbf{62.3} & \textbf{42.6} & \textbf{22.9} & \textbf{6.3} & \textbf{1.0}  \\
     20.0 & \textbf{1.720} & 93.0 & 86.6 & 77.1 & \textbf{61.6} & \textbf{42.1} & \textbf{23.4} & \textbf{6.7} & \textbf{1.2} \\
     25.0 & 1.226 & 73.2 & 64.4 & 53.9 & 42.4 & 27.4 & \textbf{14.5} & \textbf{6.5} & \textbf{1.2} \\
     30.0  & 0.846 & 44.9 & 40.1 & 33.7 & 25.1 & 17.1 & \textbf{13.6} & \textbf{10.6} & \textbf{6.9}  \\
     50.0  & 0.456 & 15.2 & 14.6 & 13.8 & 12.8 & 11.8 & 10.6 & \textbf{9.8} & \textbf{9.3} \\
    \bottomrule
    \end{tabular}
\end{table*}
    
    \begin{table}[t]
\centering
\caption{Comparison of our method to stability training \cite{li2019stab} on CIFAR-10 dataset. Each of the values indicates the fraction of test samples those have $\ell_2$ certified radius larger than the threshold specified at the top row. We set our result bold-faced whenever the value improves the baseline.}
\label{tab:cifar10}
    \begin{adjustbox}{width=1\linewidth}
    \begin{tabular}{clccccccccccc}
    \toprule
    $\sigma$ &  Models (CIFAR-10) & ACR & 0.00 & 0.25 & 0.50 & 0.75 & 1.00 & 1.25 & 1.50 & 1.75 & 2.00 & 2.25 \\ 
    \midrule
    \multirow{5}{*}{0.25}& Gaussian \cite{pmlr-v97-cohen19c} & {0.424} & {76.6} & {61.2} & {42.2} & {25.1} & 0.0 & 0.0 & 0.0 & 0.0 & 0.0 & 0.0  \\
    & \textbf{+ Consistency ($\lambda=20$)} & \textbf{0.552}  & 75.8 & \textbf{67.6} & \textbf{58.1} & \textbf{46.7} & 0.0 & 0.0 & 0.0 & 0.0 & 0.0 & 0.0  \\
    \cmidrule(l){2-2} \cmidrule(l){3-3} \cmidrule(l){4-13}
    & Stability \cite{li2019stab} ($\lambda=1$) & 0.408 & 71.6 & 57.8 & 40.7 & 27.0 & 0.0 & 0.0 & 0.0 & 0.0 & 0.0 & 0.0  \\
    & Stability \cite{li2019stab} ($\lambda=2$) & 0.421 & 72.3 & 58.0 & 43.3 & 27.3 & 0.0 & 0.0 & 0.0 & 0.0 & 0.0 & 0.0  \\ 
    & Stability \cite{li2019stab} ($\lambda=5,10,20$) & 0.102  & 10.7 & 10.7 & 10.7 & 10.7 & 0.0 & 0.0 & 0.0 & 0.0 & 0.0 & 0.0 \\  \midrule
    \multirow{5}{*}{0.50}& Gaussian \cite{pmlr-v97-cohen19c} & {0.525} & {65.7} & {54.9} & {42.8} & {32.5} & {22.0} & {14.1} & {8.3} & {3.9} & 0.0 & 0.0  \\
    & \textbf{+ Consistency ($\lambda=10$)}  & {\textbf{0.720}}  & 64.3 & \textbf{57.5} & \textbf{50.6} & \textbf{43.2} & \textbf{36.2} & \textbf{29.5} & \textbf{22.8} & \textbf{16.1} & 0.0 & 0.0  \\
    \cmidrule(l){2-2} \cmidrule(l){3-3} \cmidrule(l){4-13}
    & Stability \cite{li2019stab} ($\lambda=1$) & 0.496 & 61.1 & 51.5 & 40.9 & 29.8 & 21.1 & 14.0 & 8.3 & 3.6 & 0.0 & 0.0 \\
    & Stability \cite{li2019stab} ($\lambda=2$)  & {0.521} & 60.6 & 51.5 & 41.4 & 32.5 & 23.9 & 15.3 & 9.6 & 5.0 & 0.0 & 0.0 \\ 
    & Stability \cite{li2019stab} ($\lambda=5,10,20$) & {0.206} & 10.8 & 10.8 & 10.8 & 10.8 & 10.8 & 10.8 & 10.8 & 10.8 & 0.0 & 0.0 \\ 
    \midrule
    \multirow{5}{*}{1.00}& Gaussian \cite{pmlr-v97-cohen19c} & 0.542 & 47.2 & 39.2 & 34.0 & 27.8 & 21.6 & 17.4 & 14.0 & 11.8 & 10.0 & 7.6 \\
    & \textbf{+ Consistency ($\lambda=10$)}   & {\textbf{0.756}} & 46.3 & \textbf{42.2} & \textbf{38.1} & \textbf{34.3} & \textbf{30.0} & \textbf{26.3} & \textbf{22.9} & \textbf{19.7} & \textbf{16.6} & \textbf{13.8} \\
    \cmidrule(l){2-2} \cmidrule(l){3-3} \cmidrule(l){4-13}
    & Stability \cite{li2019stab} ($\lambda=1$) & 0.526  & 43.5 & 38.9 & 32.8 & 27.0 & 23.1 & 19.1 & 15.4 & 11.3 & 7.8 & 5.7 \\
    & Stability \cite{li2019stab} ($\lambda=2$) & 0.414 & 17.0 & 16.3 & 15.4 & 14.6 & 13.7 & 12.6 & 12.1 & 11.2 & 10.3 & 9.8  \\ 
    & Stability \cite{li2019stab} ($\lambda=5,10,20$) & 0.381 & 10.0 & 10.0 & 10.0 & 10.0 & 10.0 & 10.0 & 10.0 & 10.0 & 10.0 & 10.0  \\ 
    \bottomrule
    \end{tabular}
    \end{adjustbox}
    \vspace{-0.15in}
\end{table}
    
    \clearpage
    \section{Overview on prior works}
    
    For completeness, we present a brief introduction to the prior works mainly considered in our experiments. We use the notations defined in Section~2 of the main text throughout this section.
    
    \subsection{SmoothAdv}
    
    Recall that a smoothed classifier $\hat{f}$ is defined from a hard classifier $f:\mathbb{R}^d \rightarrow \mathcal{Y}$, namely:
    \begin{equation}
    \label{eq:smoothing}
        \hat{f}(x) := \argmax_{k\in \mathcal{Y}} \mathbb{P}_{\delta\sim\mathcal{N}(0, \sigma^2 I)}\left(f (x + \delta)= k \right).
    \end{equation}
    Here, \emph{SmoothAdv} \cite{nips_salman19} attempts to perform adversarial training \cite{madry2018towards} directly on $\hat{f}$:
    \begin{equation}
    \label{eq:smoothadv_train}
        \min_{\hat{f}} \max_{||x' - x||_2 \le \epsilon}\mathcal{L}(\hat{f}; x', y),
    \end{equation}
    where $\mathcal{L}$ denotes the standard cross-entropy loss. 
    As mentioned in the main text, however, $\hat{f}$ is practically a non-differentiable object when \eqref{eq:smoothing} is approximated via Monte Carlo sampling, making it difficult to optimize the inner maximization of \eqref{eq:smoothadv_train}. To bypass this, \citet{nips_salman19} propose to attack the \emph{soft-smoothed} classifier $\hat{F}:=\mathbb{E}_\delta[F_y(x+\delta)]$ instead of $\hat{f}$, as $\hat{F}: \mathbb{R}^d \rightarrow \Delta^{K-1}$ is rather differentiable.
    Namely, SmoothAdv finds an adversarial example via solving the following:
    \begin{equation}
    \label{eq:smoothadv_train_approx}
        \hat{x} = \argmax_{||x' - x||_2 \le \epsilon}\mathcal{L}(\hat{F}; x', y)
        = \argmax_{||x' - x||_2 \le \epsilon}\left(-\log \mathbb{E}_\delta\left[F_y(x'+\delta)\right]\right).
    \end{equation}
    
    In practice, the expectation in this objective \eqref{eq:smoothadv_train_approx} is approximated via Monte Carlo integration with $m$ samples of $\delta$, namely $\delta_1, \cdots, \delta_m \sim \mathcal{N}(0, \sigma^2 I)$:
    \begin{equation}
    \label{eq:smoothadv_train_approx2}
        \hat{x} = \argmax_{||x' - x||_2 \le \epsilon}\left(-\log \left(\frac{1}{m} \sum_i F_y(x'+\delta_i)\right)\right).
    \end{equation}
    To optimize the outer minimization objective in \eqref{eq:smoothadv_train}, on the other hand, SmoothAdv simply minimize the averaged loss over $(\hat{x}+\delta_1, y), \cdots, (\hat{x}+\delta_m, y)$, i.e., $\min_{F} \frac{1}{m}\sum_{i} \mathcal{L}(F; \hat{x}+\delta_i, y)$. Notice that the noise samples $\delta_1, \cdots, \delta_m$ are re-used for the outer minimization as well.
    
    \subsection{MACER}
    
    On the other hand, MACER \cite{Zhai2020MACER} attempts to improve robustness of $\hat{f}$ via directly maximizing the certified lower bound over $\ell_2$-adversarial perturbation \cite{pmlr-v97-cohen19c} for $(x, y)\in\mathcal{D}$: 
    \begin{equation}
        \min_{\hat{f} (x')\ne y} ||x' - x||_2 
        \ge \frac{\sigma}{2}\left(\Phi^{-1}(p^{(1)}) - \Phi^{-1}(p^{(2)}) \right),
        \label{eq:cr}
    \end{equation}
    where $p^{(1)}:=\mathbb{P}(f (x + \delta)= \hat{f}(x))$ and $p^{(2)}:=\max_{c\ne \hat{f} (x)}\mathbb{P}(f (x + \delta)= c)$, as defined in Section~2 in the main text. Again, directly maximizing \eqref{eq:cr} is difficult due to the non-differentiability of $\hat{f}$, thereby MACER instead maximizes the certified radius of $\hat{F}$, in a similar manner to SmoothAdv \cite{nips_salman19}:
    \begin{equation}
        \mathrm{CR}(\hat{F}; x, y) := \frac{\sigma}{2}\left(\Phi^{-1}(\mathbb{E}_{\delta}[F_y(x+\delta)]) - \Phi^{-1}(\max_{c\ne y}\mathbb{E}_{\delta}[F_c(x+\delta)]) \right).
        \label{eq:cr_approx}
    \end{equation}
    Motivated from the 0-1 robust classification loss \eqref{eq:decomp_01}, \citet{Zhai2020MACER} propose a robust training objective for maximizing $\mathrm{CR}(\hat{F}; x, y)$ along with the standard cross-entropy loss $\mathcal{L}$ on $\hat{F}$ as a surrogate loss for the natural error term:
    \begin{equation}
        L_\varepsilon(f) :=\ \mathbb{E}_{(x, y)\in\mathcal{D}}\left[1 - \mathbf{1}_{\mathrm{CR}(\hat{f}; x, y) \ge \varepsilon}\right]
        = \underbrace{\mathbb{E}\left[\mathbf{1}_{\hat{f}(x) \ne y}\right]}_{\text{natural error}} + \underbrace{\mathbb{E}\left[\mathbf{1}_{\hat{f}(x) = y,\ \mathrm{CR}(\hat{f}; x, y) < \varepsilon}\right]}_{\text{robust error}}
        \label{eq:decomp_01}
    \end{equation}
    \begin{equation}
        L_{\tt MACER}(F; x, y) :=
        \underbrace{\mathcal{L}(\hat{F}(x), y)}_{\text{natural error}} +\ \lambda \cdot \underbrace{\frac{\sigma}{2}\max\{\gamma-\mathrm{CR}(\hat{F}; x, y), 0\}\cdot \mathbf{1}_{\hat{F}(x)=y}}_{\text{robust error}},
        \label{eq:decomp_macer}
    \end{equation}
    where $\gamma$, $\lambda$ are hyperparameters. Here, notice that \eqref{eq:decomp_macer} uses the hinge loss to maximize $\mathrm{CR}(\hat{F}; x, y)$, only for the samples that $\hat{F}(x)$ is correctly classified to $y$. 
    In addition, MACER uses an inverse temperature $\beta > 1$ to calibrate $\hat{F}$ as another hyperparameter, mainly for reducing the practical gap between $\hat{F}$ and $\hat{f}$.
    
    \clearpage
    \small
    \bibliographystyle{plainnat}
    \bibliography{references}